\documentclass[hidelinks,onefignum,onetabnum]{siamart250211}
\usepackage{amsmath,amsfonts}
\usepackage{algorithmic}
\usepackage{algorithm}
\usepackage{array}
\usepackage[caption=false,font=normalsize,labelfont=sf,textfont=sf]{subfig}
\usepackage{textcomp}
\usepackage{stfloats}
\usepackage{url}
\usepackage{verbatim}
\usepackage{graphicx}
\usepackage{cite}
\usepackage{makecell}
\usepackage{multirow}
\usepackage{tabularx,booktabs}
\usepackage{caption}
\usepackage{amssymb}
\usepackage[figuresright]{rotating}
\usepackage{tikz}
\usepackage{microtype}

\usepackage{lipsum}
\usepackage{amsfonts}
\usepackage{graphicx}
\usepackage{epstopdf}
\usepackage{algorithmic}
\ifpdf
  \DeclareGraphicsExtensions{.eps,.pdf,.png,.jpg}
\else
  \DeclareGraphicsExtensions{.eps}
\fi

\newsiamremark{remark}{Remark}
\newsiamremark{hypothesis}{Hypothesis}
\crefname{hypothesis}{Hypothesis}{Hypotheses}
\newsiamthm{claim}{Claim}
\newsiamremark{fact}{Fact}
\crefname{fact}{Fact}{Facts}

\headers{Hyperbolic Large Language Models (LLMs)}{S. Patil, Z. Zhang, Y. Huang, T. Ma, and M. Xu}

\title{Hyperbolic Large Language Models \thanks{
\funding{This work was supported by the DOE SEA-CROGS project (DE-SC0023191), AFOSR project (FA9550-24-1-0231). (\it Corresponding author:} \email{mx6@njit.edu})}}

\newcommand{\affilSBUBI}{Department of Biomedical Informatics, Stony Brook University, NY, USA}

\author{
Sarang Patil\thanks{Department of Data Science, New Jersey Institute of Technology, Newark, NJ, USA (\email{sp3463@njit.edu}, \email{zz63@njit.edu}, \email{yh87@njit.edu}).}
\and Zeyong Zhang\footnotemark[2]
\and Yiran Huang\footnotemark[2]
\and Tengfei Ma\thanks{\affilSBUBI (\email{tengfei.ma@stonybrook.edu}).} 
\and Mengjia Xu\footnotemark[2]
}

\usepackage{amsopn}

\ifpdf
\hypersetup{
  pdftitle={Hyperbolic Large Language Models},
  pdfauthor={S. Patil, Z. Zhang, Y. Huang, T. Ma and M. Xu}
}
\fi

\begin{document}

\maketitle

\begin{abstract}
Large language models (LLMs) have achieved remarkable success and demonstrated superior performance across various tasks, including natural language processing (NLP), weather forecasting, biological protein folding, text generation, and solving mathematical problems. However, many real-world data exhibit highly non-Euclidean latent hierarchical anatomy, such as protein networks, transportation networks, financial networks, brain networks, and linguistic structures or syntactic trees in natural languages. Effectively learning intrinsic semantic entailment and hierarchical relationships from these raw, unstructured input data using LLMs remains an underexplored area. Due to its effectiveness in modeling tree-like hierarchical structures, hyperbolic geometry -- a non-Euclidean space -- has rapidly gained popularity as an expressive latent representation space for complex data modeling across domains such as graphs, images, languages, and multi-modal data. Here, we provide a comprehensive and contextual exposition of recent advancements in LLMs that leverage hyperbolic geometry as a representation space to enhance semantic representation learning and multi-scale reasoning. Specifically, the paper presents a taxonomy of the principal techniques of Hyperbolic LLMs (HypLLMs) in terms of four main categories: (1) hyperbolic LLMs through exp/log maps; (2) hyperbolic fine-tuned models; (3) fully hyperbolic LLMs, and (4) hyperbolic state-space models. We also explore crucial potential applications and outline future research directions. A repository of key papers, models, datasets, and code implementations is available at ~\url{https://github.com/sarangp2402/Hyperbolic-LLM-Models}.
\end{abstract}



\section{Introduction}
The emergence of foundation models and large language models (LLMs) has significantly empowered modern artificial intelligence (AI) systems, producing substantial impacts across a wide range of application fields, such as healthcare~\cite{kather2024large, abramson2024accurate}, finance~\cite{li2023large}, computer programming~\cite{chen2021evaluating, wang2024enhancing}, earth system modeling~\cite{bodnar2024aurora,thulke2024climategpt}, autonomous systems~\cite{xia2023towards}, etc. Prevalent LLMs, such as GPT~\cite{achiam2023gpt}, Llama~\cite{dubey2024llama}, Gemini~\cite{team2023gemini}, Claude~\cite{TheC3}, BERT~\cite{kenton2019bert}, built on transformer architectures~\cite{vaswani2017attention} perform effectively in various natural language processing (NLP) tasks, such as summarization, answering questions, translation, and generation of coherent and contextually relevant text. Typically, LLMs were designed to learn token embeddings in flat Euclidean space, where each token is represented as a single vector, leveraging the attention mechanism to capture rich contextual information. However, Euclidean embeddings present two major limitations: 1) flat Euclidean space (with curvature $\mathcal{K} = 0$) has limited capacity to embed large and complex taxonomies with high distortions; 2) they struggle to capture logical and semantic entailment information during representation learning, including latent hierarchical structural features, which are critical for complex semantic reasoning and question answering tasks. This limitation is particularly evident when processing complex multi-modal data that exhibits tree-like structures with multi-scale features at various levels. Furthermore, numerous real-world datasets inherently exhibit hierarchical tree-like structures in non-Euclidean space~\cite{nickel2017poincare, krioukov2010hyperbolic}, such as textual entailment ontology data, linguistic structures, and phylogenetic trees of DNA sequences~\cite{boguna2021network}. However, hyperbolic space with negative curvature ($\mathcal{K} < 0$) is well-suited for representing the aforementioned large-scale hierarchical datasets, offering significant advantages over conventional Euclidean space for three key properties: 1) Hyperbolic space grows exponentially, aligning naturally with the tree-like expansion of hierarchical data, allowing for the effective preservation and representation of hierarchies with lower distortion. 2) Hyperbolic embeddings excel at preserving both local (leaf-level) and global (root-level) relationships, ensuring a comprehensive representation of hierarchical patterns. 3) Hyperbolic space offers high representational capacity even in low dimensions, reducing computational overhead.

Learning high-quality representations in hyperbolic space has gained growing interest in graph machine learning, computer vision, and natural language processing. In geometric deep learning, many real-world networks are scale-free, motivating the development of hyperbolic extensions of traditional GNNs, such as hyperbolic skip-gram models~\cite{chamberlain2017neural}, hyperbolic GNNs~\cite{liu2019hyperbolic}, hyperbolic graph convolutional networks (GCNs)~\cite{chami2019hyperbolic}, and fully hyperbolic GNNs~\cite{wang2021fully}. These models effectively capture hierarchical structures and large-scale graph geometry with fewer parameters and stronger generalization, showing superior performance in tasks like knowledge graph completion~\cite{krioukov2010hyperbolic, kolyvakis2019hyperkg, chami2020low, wang2021knowledge, xiao2022complex}, and brain network embedding for disease detection and brain age prediction~\cite{baker2024hyperbolic, ramirez2024fully}. 
Previous works~\cite{yang2022hyperbolic, peng2021hyperbolic} provide a comprehensive review of hyperbolic deep neural networks or hyperbolic graph neural network methods and their applications across multiple domains. However, despite these advances in hyperbolic embeddings for specific domains, a comprehensive understanding of how hyperbolic geometry can be systematically integrated into large language models remains fragmented across the literature.

For natural language data, the word frequency typically follows a Zipfian distribution, as described by Zipf’s law~\cite{piantadosi2014zipf}, which reflects its hierarchical structure and motivates the use of hyperbolic geometry for representation. Petrovski~\cite{petrovski2024hyperbolic} demonstrated that embedding 
sentences in the Poincar\'e disk using Möbius averaging outperforms Euclidean averaging on textual-entailment benchmarks, showing that negative curvature can benefit natural-language inference. To capture useful word-level hierarchies for a variety of NLP tasks, hyperbolic geometry (i.e., Poincar\'e ball model) has been used to enhance conventional word embedding methods (e.g., word2vec and GloVe)~\cite{nickel2017poincare,tifrea2019poincar}. In particular, hyperbolic word embeddings outperform their Euclidean counterparts in lexical entailment prediction, offering higher efficiency and better generalization.
They also achieve exceptional performance in word similarity, analogy, and hypernymy detection using Gaussian mappings on the Riemannian manifold~\cite{tifrea2019poincar}. Several architectures extend RNNs and GRUs into hyperbolic space via gyrovector operations or Lorentz transformations, enabling end-to-end training without returning to tangent space~\cite{ganea2018HNNs, chen2021fully}. The results demonstrate significant improvements in the performance of natural language inference and noisy-prefix recognition. Fully hyperbolic neural networks~\cite{chen2021fully} for NLP tasks represent a significant advancement by operating entirely within hyperbolic space using Lorentz transformations, overcoming the limitations of previous hybrid approaches that relied on tangent space operations. 

More recently, pre-trained LLMs have been found  exhibit intrinsic hyperbolic structure, with token frequencies following a power law ($\alpha \approx 1.9$) and embeddings showing tree-like organization ($\delta$-hyperbolicity $\approx 0.08-0.12$)~\cite{yang2024hyperbolic}. These findings provide compelling empirical evidence that hyperbolic geometry aligns naturally with the intrinsic structure of language model representations, validating the theoretical foundations for hyperbolic LLMs (HypLLMs) development. Herein, we primarily focus on recent advances in developing powerful LLMs that leverage hyperbolic geometry as their representation space, aiming to further enhance expressiveness, improve reasoning, capture hierarchical structures, and optimize efficiency for large-scale complex data. Note that we use the term ``large language models'' (LLMs) broadly to encompass neural network architectures, primarily but not exclusively based on transformers, that are trained on extensive datasets to achieve strong performance across different tasks. While initially developed for natural language processing, modern LLMs include foundation models operating on diverse modalities such as vision, audio, and multimodal data. Modern LLMs operate over vocabularies on the order of 50K-100K+ tokens~\cite{takase2024large}, yet the complex semantic hierarchies within the data is not well-captured by Euclidean embeddings, which have reduced representation capacity for hierarchical data compared to hyperbolic alternatives. While the dominant paradigm has been Transformer-based architectures, which excel as LLMs, the field has also witnessed the emergence of alternative architectures such as state-space models (SSMs), with Mamba being a prominent example that offers improved computational efficiency over long-range dependencies~\cite{gu2023mamba}. Mamba's improved successor Mamba2~\cite{dao2024transformers} has also shown strong performance implementing state-space duality between attention and state space modeling. Both Transformer and Mamba architectures present unique opportunities and challenges when adapted to hyperbolic geometry, which we explore throughout this paper. 
This paper serves as a comprehensive survey that synthesizes and unifies the fragmented landscape of hyperbolic geometry integration in large language models. Our primary goal is to provide a coherent mathematical and architectural framework of existing approaches, clarifying their relationships, and establishing a foundation for future research in this emerging field. In Section~\ref{sec:foundations}, we first present the foundations of hyperbolic geometry, representation models, manifold mappings, and key optimization techniques in the hyperbolic manifold. Section~\ref{sec:methods} presents a taxonomy of existing hyperbolic LLM approaches. Section~\ref{sec:benchmarking} provides a benchmarking analysis of hyperbolic LLMs, evaluating their performance on mathematical reasoning and hierarchical language reasoning, while identifying current benchmarking limitations and the need for standardized evaluation frameworks. We also discuss the key challenges in building hyperbolic LLMs in this section. Section~\ref{sec:applications} explores emerging applications of hyperbolic LLMs across three major domains: natural language processing, computer vision, and multi-modal visual-semantic representation and inference. Finally, we conclude the paper with a discussion on future research directions.

\section{Foundations of hyperbolic geometry}
\label{sec:foundations}

\begin{figure}[ht]
\centering
\begin{tikzpicture}
  \matrix[column sep=0.3cm, row sep=2mm]{ 
    \node (euclPic) {
      \begin{tikzpicture}[scale=0.7] 
        \draw[thick, blue] (-1.2,0) -- (1.2,0);
      \end{tikzpicture}
    }; &
    \node (sphPic) {
      \begin{tikzpicture}[scale=0.7] 
        \draw[thick, orange] (-1.0,0) arc (180:0:1.0);
      \end{tikzpicture}
    }; &
    \node (hypPic) {
      \begin{tikzpicture}[scale=0.7] 
        \draw[thick, red] (-1.2,0.08) .. controls (-0.6,-0.28) and (0.6,-0.28) .. (1.2,0.08);
        \draw[thick, red] (-0.8,0.36) .. controls (-0.32,-0.04) and (0.32,-0.04) .. (0.8,0.36);
      \end{tikzpicture}
    }; \\
    \node[text width=2.6cm, align=center] { 
      \small\textbf{Euclidean ($\mathcal{K}=0$)}\\[0.5mm] 
      \footnotesize \emph{Can be used for generic data}\\[-0.5mm] 
    }; &
    \node[text width=2.6cm, align=center] { 
      \small\textbf{Spherical ($\mathcal{K}>0$)}\\[0.5mm] 
      \footnotesize \emph{Can be used for globally compact, closed topology}\\[-0.5mm] 
    }; &
    \node[text width=2.6cm, align=center] { 
      \small\textbf{Hyperbolic ($\mathcal{K}<0$)}\\[0.5mm] 
      \footnotesize \emph{Can be used for hierarchical, tree structures}\\[-0.5mm] 
    }; \\
  };
\end{tikzpicture}
\caption{Three commonly used representation spaces: Euclidean space, where traditional LLMs operate; spherical geometry, which constrains embeddings to bounded surfaces; and hyperbolic space, the basis of Hyperbolic LLMs, which naturally models hierarchical structures.}
\label{fig:curvature_spaces}
\vspace{-.18in}
\end{figure}
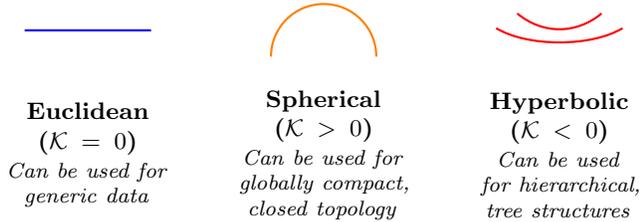

Hyperbolic space $\mathbb{H}^n$ is a non-Euclidean geometric space with constant sectional negative curvature ($\mathcal{K}<0$), where the parallel postulate fails, i.e., allowing multiple parallel lines to pass through a point outside a given line. It can be viewed as a ``continuous version of trees'', characterized by negative curvature that makes distances grow exponentially toward the boundary. Notably, hyperbolic space shares a deep connection with Minkowski spacetime in special relativity, as the Lorentz model of hyperbolic geometry is isometric to the hyperboloid model used in Minkowski space~\cite{barrett2011hyperbolic, nickel2018learning}. 
The three fundamental geometric spaces that underpin different representation learning approaches are Euclidean ($\mathcal{K}=0$, traditional LLMs), spherical ($\mathcal{K}>0$, compact topology), and hyperbolic ($\mathcal{K}<0$, hierarchical data) as represented in \Cref{fig:curvature_spaces}. 
Moreover, hyperbolic embeddings can represent hierarchical structures much more parsimoniously – for instance, Nickel and Kiela~\cite{nickel2017poincare} showed that a 5-dimensional hyperbolic model captured the WordNet noun hierarchy better than a 200-dimensional Euclidean model, highlighting the vastly greater representational capacity of negatively curved space.

\subsection{Hyperbolic representation models}
\begin{figure}[ht]
\centering
\includegraphics[width=3.5in]{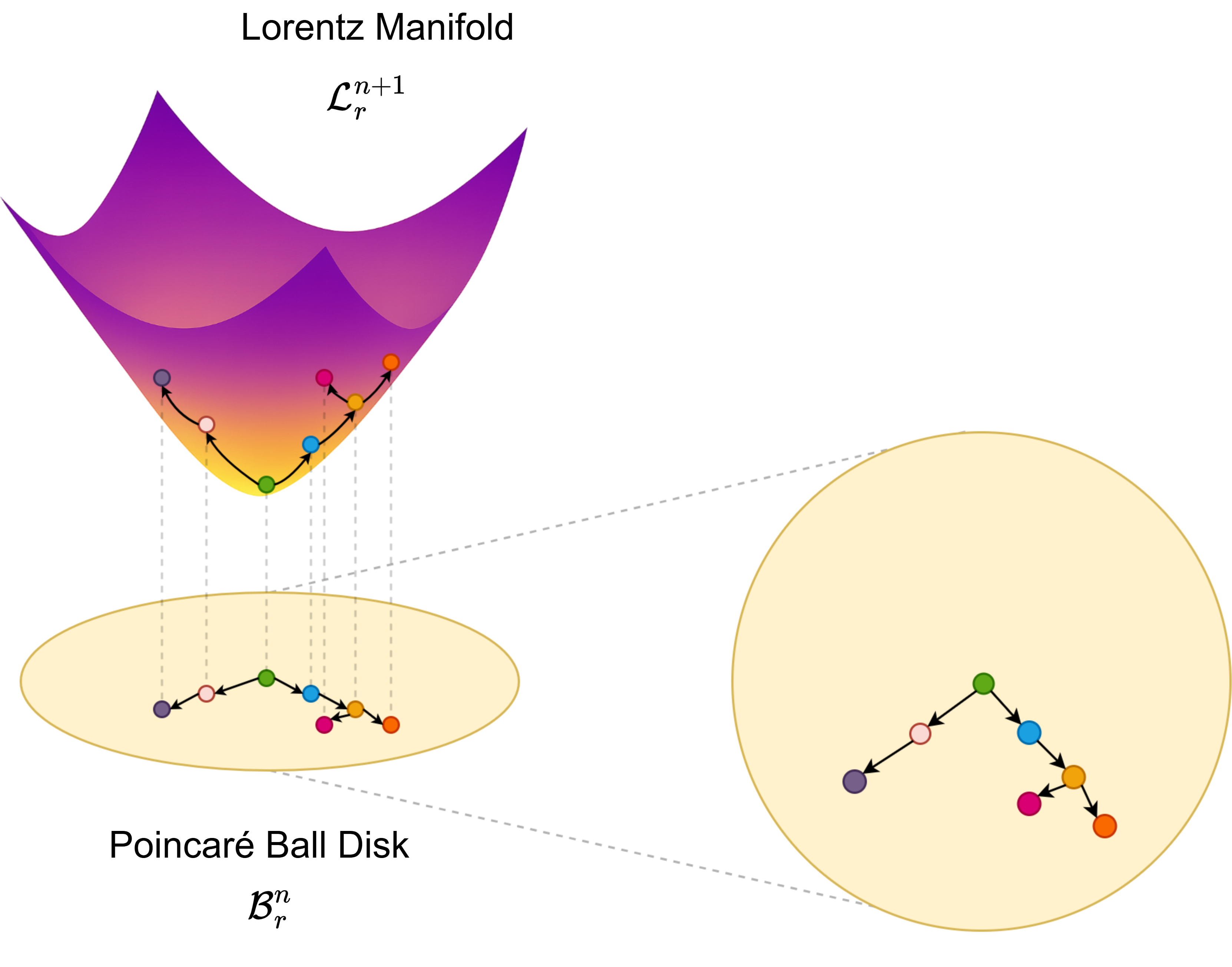}
 \caption{Illustration of the two primary hyperbolic representation models used in HypLLMs: the \textbf{Lorentz model} (top left) and the \textbf{Poincar\'e ball model} (right). The Lorentz model represents hyperbolic space on a single sheet of hyperboloid embedded in Minkowski space $\mathcal{L}_r^{n+1}$, offering closed-form geodesic computations and improved numerical stability. The lower oval shows Poincar\'e ball disk projection in space $\mathcal{B}_r^n$,  this a conformal model where geodesics are circular arcs and points near the boundary encode fine-grained leaf-level hierarchy. These geometric representations provide the foundation for hierarchical embeddings in hyperbolic LLMs.}
\label{fig:hyperbolic_geometry}
\vspace{-.32in}
\end{figure}

Two commonly used isomorphic models for representing hyperbolic space, the Poincar\'e ball and the Lorentz model (or hyperboloid manifold), are described in the \Cref{fig:hyperbolic_geometry}. The sectional curvature of hyperbolic space is determined by $\mathcal{K} = -1/r^2$, where $r > 0$ is the radius that determines the amount of curvature, i.e., {\it larger $r$ $\rightarrow$ smaller curvature ($\mathcal{K}$) $\rightarrow$ flatter space}; $\mathcal{K} = 0 $ is Euclidean space when $r\rightarrow\infty$. In the standard hyperbolic space with sectional curvature $\mathcal{K} = -1$, the curvature parameter $r = 1$. Both $\mathcal{B}_r^n$ and $\mathcal{L}_r^{n+1}$ represent the same abstract $n$-dimensional hyperbolic space $\mathbb{H}^n$ with radius $r$, but use different coordinate systems. The Poincar\'e ball $\mathcal{B}_r^n$ provides a bounded $n$-dimensional representation, while the Lorentz model $\mathcal{L}_r^{n+1}$ embeds the hyperbolic space as a hyperboloid in $(n+1)$-dimensional Minkowski spacetime.

\subsubsection*{(1) Poincar\'e ball model}
An $n$-dimensional \emph{Poincar\'e ball} model represents $\mathbb{H}^n$ as the open unit ball $ \mathcal{B}_r^n = \{x \in \mathbb{R}^n: \|x\| < r\}$ (where $\|\cdot\|$ denotes the Euclidean norm) with metric $ds^2 = \lambda_x^2\,dx^2$, where $dx^2$ is the Euclidean metric and $\lambda_x = \frac{2}{1-\|x\|^2}$ is the conformal factor that scales distances according to the hyperbolic geometry~\cite{nickel2017poincare}. Geodesics in this manifold are arcs that tend toward the boundary $\|x\|\to 1$. The geodesic distance between two points $u,v \in \mathcal{B}_r^n$ is:
\begin{equation}
d_{\mathcal{B}}(u,v) = r \cdot \mathrm{arcosh}\!\Big(1 + 2\,\frac{\|u-v\|^2}{(r^2-\|u\|^2)(r^2-\|v\|^2)}\Big),
\end{equation}

which grows without bound as $u$ or $v$ approach the boundary of the ball. 
This model is advantageous for gradient-based optimization.

\subsubsection*{(2) Lorentz model}
Alternatively, the \emph{Lorentz} model represents $\mathbb{H}^n$ as a hyperboloid embedded in $\mathbb{R}^{n+1}$: $\mathcal{L}_r^{n+1} = \{x \in \mathbb{R}^{n+1}: \langle x,x\rangle_L = -r^2,\; x_0>0\}$, equipped with the Minkowski inner product~\cite{nickel2018learning}:
\begin{equation}
\label{eq:minkowski}
\langle x,y\rangle_L = -x_0y_0 + \sum_{i=1}^{n}x_i y_i.
\end{equation}

Note that the Lorentz model uses only the \textit{upper sheet} ($x_0 > 0$) of the two-sheeted hyperboloid. This upper sheet constraint ensures that all points represent valid hyperbolic space coordinates. The hyperbolic distance between two points $x$ and $y$ is:
\begin{equation}
\label{eq:lorentz}
d_{\mathcal{L}}(x,y) = r \cdot \operatorname{arcosh}\left(\frac{-\langle x,y\rangle_L}{r^2}\right),
\end{equation}
and one can convert between the Lorentz and Poincar\'e representations via explicit coordinate mappings~\cite{pmlr-v202-mishne23a}. Notably, the Lorentz model often provides better numerical stability for optimization. In either models, gradient-based learning can be performed with Riemannian optimization techniques in Section~\ref{sec:hyp_optimizers} and projected Euclidean updates via the exponential and logarithmic maps.

\subsection{Exponential and logarithmic mapping}

\begin{figure}[ht]
\centering
\includegraphics[width=2.7in]{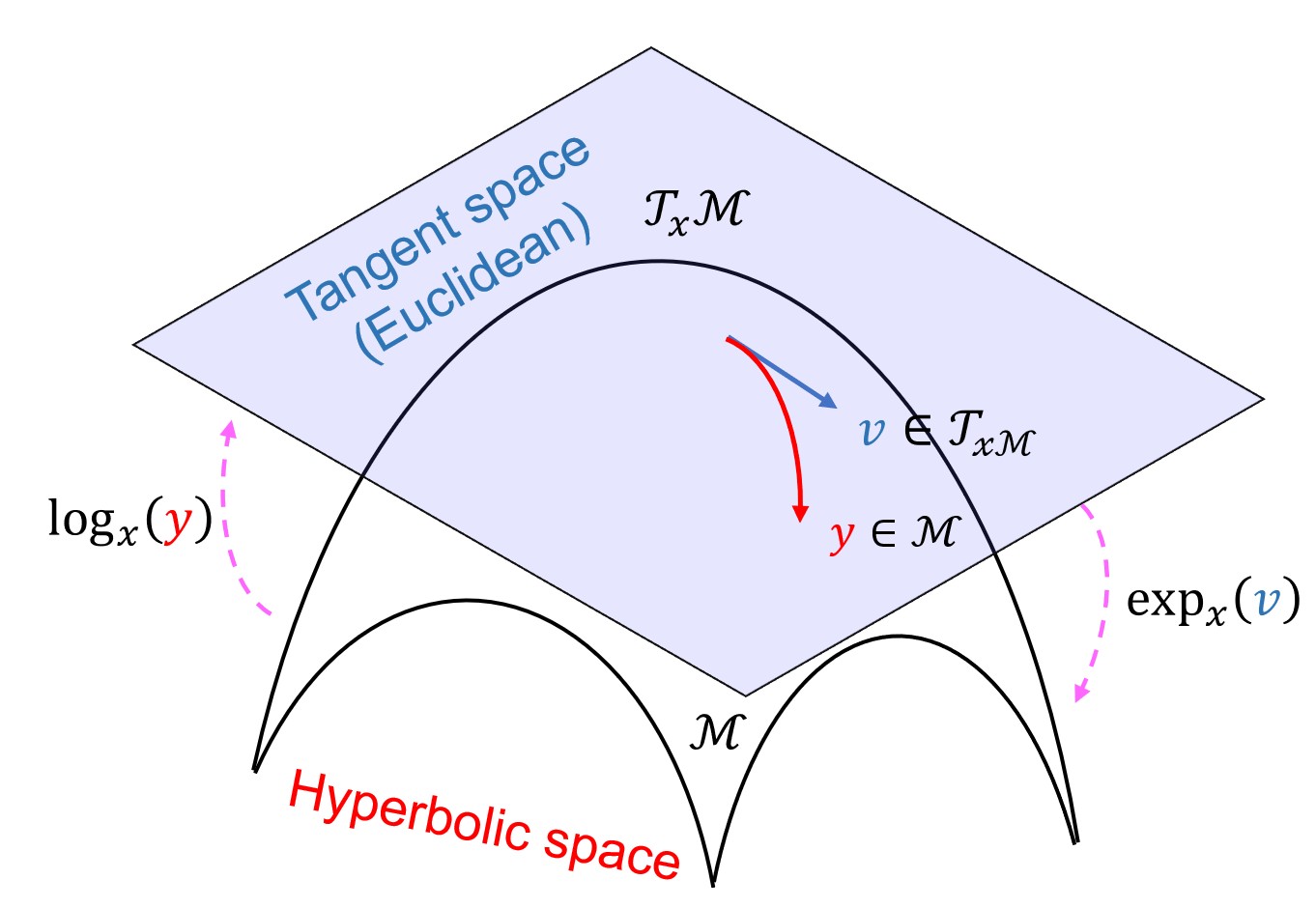}
\caption{Illustration of exponential and logarithmic mappings between Manifold $\mathcal{M}$ (Poincar\'e ball or Lorentz model) and the tangent space $\mathcal{T}_x\mathcal{M}$ (Euclidean). 
The exponential map $\exp_x(v)$ projects a tangent vector $v \in \mathcal{T}_x\mathcal{M}$ onto the manifold, while the logarithmic map $\log_x(y)$ projects a manifold point $y \in \mathcal{M}$ back to the tangent space.}
\label{fig:tangent_space_mapping}
\vspace{-.25in}
\end{figure}

Given a hyperbolic manifold $\mathcal{M}$ (either the Poincar\'e ball or Lorentz model in $\mathbb{H}^n$) and a point $x \in \mathcal{M}$, the tangent space $\mathcal{T}_x\mathcal{M}$ corresponds to $\mathbb{R}^n$. The exponential map $\exp_x : \mathcal{T}_x\mathcal{M} \to \mathcal{M}$ maps a tangent vector at $x$ to the point on the manifold $\mathcal{M}$ reached by following the geodesic starting at $x$. The logarithmic map $\log_x : \mathcal{M} \to \mathcal{T}_x\mathcal{M}$ maps a manifold point to the tangent space at $x$ along the geodesic connecting it to $x$. These bidirectional maps connect the hyperbolic manifold $\mathcal{M}$ and its tangent space $\mathcal{T}_x\mathcal{M}$, as shown in Figure~\ref{fig:tangent_space_mapping}.

(1) For the Poincar\'e Ball Model ($\mathcal{M}=\mathcal{B}_r^n$), the exponential map of vector $v \in \mathcal{T}_x\mathcal{M}$ to the manifold $\mathcal{M}$ at any point $x\in \mathcal{M}$ is given by,
\begin{equation}
\text{exp}_x(v) = x \oplus_r 
\left(\tanh\left(\frac{\lambda_x\ \|v\|}{2r}\right) 
\frac{r v}{\|v\|}\right),
\end{equation}
where $\lambda_x$ is the conformal factor and $\oplus_r$ denotes Möbius addition. The logarithmic map of $y \in \mathcal{M}$ to the tangent space $\mathcal{T}_x\mathcal{M}$ is given by,
\begin{equation}
\text{log}_x(y) = \frac{2r}{\lambda_x} 
\text{artanh}\left(\frac{\|{-}x \oplus_r y\|}{r}\right) 
\frac{{-}x \oplus_r y}{\|{-}x \oplus_r y\|}.
\end{equation}

(2) For the Lorentz model ($\mathcal{M}=\mathcal{L}^{n+1}_r$), the exponential and logarithmic maps at any point $x \in \mathcal{M}$ are defined as follows~\cite{chami2019hyperbolic},
\begin{equation}
\text{exp}_x(v) = \cosh\left(\frac{\sqrt{\langle v, v\rangle_{\mathcal{L}}}}{r}\right) 
x + r\sinh\left(\frac{\sqrt{\langle v,v \rangle_{\mathcal{L}}}}{r}\right) 
\frac{v}{\sqrt{\langle v, v\rangle_{\mathcal{L}}}},
\end{equation}
\begin{equation}
\text{log}_{x}(y) = d_{\mathcal{L}}(x,y)
\frac{y + (r^{-2} \langle x, y\rangle_{\mathcal{L}}) x}
{\|y + (r^{-2}\langle x, y\rangle_{\mathcal{L}}) x\|_\mathcal{L}},
\end{equation}
where $d_{\mathcal{L}}(x,y)$ is the hyperbolic distance defined in Equation~\ref{eq:lorentz}. These mappings preserve radial distances and enable a hybrid workflow: (i) hyperbolic points are mapped to the tangent space via $\text{log}_x$, (ii) standard Euclidean operations are performed, and (iii) results are mapped back to hyperbolic space via $\text{exp}_x$. This bidirectional mapping between the hyperbolic manifold $\mathcal{M}$ and its tangent space $\mathcal{T}_x\mathcal{M}$ enables hybrid models to combine Euclidean computational efficiency with hyperbolic geometric fidelity and hierarchical expressiveness.

To enable direct computation in hyperbolic space, Ungar~\cite{ungar2008analytic, ungar2022gyrovector} introduced Möbius gyrovector algebra, enabling vector space operations in the Poincar\'e ball~\cite{ganea2018HNNs} (see Appendix~\ref{sec:mobius}). These operations preserve the hyperbolic structure while enabling neural network computations such as linear transformations and activations.

\subsection{Optimization techniques in Riemannian manifold}\label{sec:hyp_optimizers}
Training models in hyperbolic space requires specialized optimization techniques, since Euclidean gradient descent is incompatible with curved manifolds. Gradients are computed in the tangent space and then mapped back to the manifold, allowing intrinsic parameter updates~\cite{bonnabel2013stochastic}.
A key insight is to perform gradient updates intrinsically on the manifold via \textit{Riemannian optimization}~\cite{liu2018Riemannian}.

\textbf{Riemannian Stochastic Gradient Descent (RSGD)}: RSGD generalizes SGD to Riemannian manifolds by using exponential mappings to update parameters while maintaining manifold constraints~\cite{bonnabel2013stochastic}:
\vspace{-0.1cm}
\begin{equation}
\theta_{t+1} = \exp_{\theta_t}(-\eta_t \nabla_R \mathcal{L}(\theta_t)),
\end{equation}
where $\exp_{\theta_t}$ is the exponential map at parameter $\theta_t$ (point on the manifold), $\eta_t$ is the learning rate, and $\nabla_R \mathcal{L}(\theta_t)$ is the Riemannian gradient of the loss function. Bonnabel first proved that RSGD updates converge under certain conditions on geodesically convex functions, establishing the theoretical foundation for hyperbolic optimization~\cite{bonnabel2013stochastic}.

\textbf{Riemannian Stochastic Variance Reduced Gradient (RSVRG)}: Building on this foundation, RSVRG extends the Stochastic Variance Reduced Gradient (SVRG) method to non-Euclidean settings to accelerate convergence. It integrates geometric principles with variance reduction, achieving linear convergence for geodesically convex functions and $\mathcal{O}(1/T)$ convergence for non-convex objectives~\cite{NIPS2016_98e6f172, liu2018Riemannian}.

\textbf{Riemannian AdaGrad:} Adaptive gradient methods have been extended to Riemannian manifolds~\cite{becigneul2018riemannian}. For a product manifold $\mathcal{M} = \mathcal{M}_1 \times \cdots \times \mathcal{M}_n$, the parameter update becomes
\vspace{-0.3cm}
\begin{equation}
\theta_{t+1}^{(i)} = \exp_{\theta_t^{(i)}}\left(-\eta_t \frac{\nabla_R^{(i)}\mathcal{L}(\theta_t)}{\sqrt{v_t^{(i)} + \epsilon}}\right),
\end{equation}
where $v_t^{(i)}$ tracks historical gradients for component $\mathcal{M}_i$, and $\epsilon$ is a small positive constant added for numerical stability to prevent division by zero. This adaptive approach demonstrates superior convergence properties compared to RSGD~\cite{becigneul2018riemannian}. These techniques significantly reduce computational overhead while maintaining the benefits of hyperbolic geometry.

In practice, modern hyperbolic LLM frameworks leverage Riemannian Adam or Riemannian AdaGrad~\cite{becigneul2018riemannian} (extensions of popular Euclidean optimizers adapted to hyperbolic manifolds) to achieve stable training. The choice between the Poincar\'e ball and Lorentz models significantly impacts optimization stability. The Lorentz model enables more stable optimization through efficient closed-form geodesic computations, avoiding numerical instabilities from Poincar\'e distance calculations while preserving the exponential growth properties essential for hierarchical representations. Critical implementation details include gradient rescaling to prevent boundary violations and appropriate initialization strategies to maintain numerical stability throughout training~\cite{pmlr-v202-mishne23a, yu2021representing}. Empirically, Nickel and Kiela~\cite{nickel2017poincare, nickel2018learning} noted that using the manifold’s geometry during optimization yields higher-quality embeddings and faster convergence than naive Euclidean training with projection. Efficient training of hyperbolic neural networks remains an active area of research, with recent advances such as Sparse Spectral Training (SST)~\cite{zhao2024sparse} offering scalable and memory-efficient optimization strategies for both Euclidean and hyperbolic models, including applications in node classification and link prediction.

\section{Hyperbolic LLMs}
\label{sec:methods}
Our taxonomy organizes the Hyperbolic LLMs into four categories (\cref{tab:hypLLM_categorized}): hybrid hyperbolic-Euclidean models, hyperbolic fine-tuned models, fully hyperbolic models, and hyperbolic state-space models. A summary of HypLLM models, including their tasks, data types, and benchmark datasets across various application domains, is provided in Appendix Table~\ref{tab:hypLLM_models_task_data}.
\begin{table*}[ht]
\vspace{-.12in}
\centering
\setlength{\tabcolsep}{3pt}
\footnotesize
\caption{Taxonomy of hyperbolic large language models (HypLLMs), organized by architecture type, backbone, model name, and geometry.}
\begin{tabular}{|l|l|l|l|}
\hline
\textbf{Category} & \textbf{Backbone} & \textbf{Model} & \textbf{Geometry} \\
\hline
\multirow{10}{*}{\makecell[l]{Hybrid hyperbolic-\\Euclidean models}}
& Transformer & Hyperbolic BERT~\cite{chen2024hyperbolic} & Lorentz \\
\cline{2-4}
& Transformer & HiT~\cite{he2024language} & Poincar\'e ball \\
\cline{2-4}
& Transformer & PoinCLIP~\cite{Srivastava2024} & Poincar\'e ball \\
\cline{2-4}
& Transformer & Hyperbolic BLIP-2~\cite{mandica2024hyperbolic} & Poincar\'e ball \\
\cline{2-4}
\cline{2-4}
& Transformer & HyperLLM~\cite{cheng2025large} & Poincar\'e ball and Lorentz \\
\cline{2-4}
& Transformer & HERec~\cite{ma2024harec} & Lorentz \\
\cline{2-4}
& Tri-Modal MoE & HyperSurv~\cite{xiongenhancing} & Lorentz \\
\cline{2-4}
& Attention-based NN & HySurvPred~\cite{yang2025hysurvpred} & Poincar\'e ball \\
\cline{2-4}
& Attention-based NN & ANTHEM~\cite{choudhary2022anthem} & Lorentz \\
\hline
\multirow{2}{*}{\makecell[l]{Hyperbolic\\fine-tuned models}}
& Transformer & HypLoRA~\cite{yang2024hyperbolic} & Lorentz \\
\cline{2-4}
& Transformer & HoRA~\cite{yang2024enhancing} & Lorentz \\
\hline
\multirow{4}{*}{\makecell[l]{Fully hyperbolic\\models}}
& Transformer & Hypformer~\cite{yang2024hypformer} & Lorentz \\
\cline{2-4}
& Transformer & HELM~\cite{he2025helm} & Lorentz \\
\cline{2-4}
& Transformer & L-CLIP~\cite{he2025hypercore} & Lorentz \\
\cline{2-4}
& GNN+Transformer & HypGraphRAG~\cite{he2025hypercore} & Lorentz \\
\hline
\multirow{5}{*}{\makecell[l]{Hyperbolic\\state-space models}}
& Mamba & SHMamba~\cite{yang2024shmamba} & Poincar\'e ball \\
\cline{2-4}
& Mamba & HMamba-Full~\cite{zhang2025hmamba} & Lorentz \\
\cline{2-4}
& Mamba & HMamba-Half~\cite{zhang2025hmamba} & Half Euclidean - Half Lorentz \\
\cline{2-4}
& Mamba-2 & HiM-Poincar\'e~\cite{patil2025hierarchical} & Poincar\'e ball \\
\cline{2-4}
& Mamba-2 & HiM-Lorentz~\cite{patil2025hierarchical} & Lorentz \\
\hline
\end{tabular}
\label{tab:hypLLM_categorized}
\end{table*}

Now we address each category with the fundamental challenges of either adapting Euclidean neural operations to curved manifolds or direct embedding in the hyperbolic manifold through distinct strategies such as hybrid models maintaining compatibility through strategic mapping, fine-tuning methods adapting existing models with minimal architectural changes, or fully hyperbolic approaches redesigning all components for native curved space operation. Given input embeddings $X \in \mathbb{R}^{B \times L \times d}$, where $B$ is the batch size, $L$ is the sequence length, and $d$ is the embedding dimension, the model outputs embeddings $Y \in \mathbb{R}^{B \times L \times d}$. Individual tokens $\mathbf{x}_1, \ldots, \mathbf{x}_L$ represent sequence positions. We denote the hyperbolic embedding space as $\mathbb{H}^n$ with Euclidean space as $\mathbb{R}^n$. We differentiate the embedding dimensions between $\mathbb{R}^n$ and $\mathbb{H}^n$ by using $d_e$ (Euclidean embedding dimensions) and $d_h$ (Hyperbolic embedding dimensions). These notations will be used throughout Figures \ref{fig:hybrid_architecture}-\ref{fig:fully_hyperbolic_architecture} to describe input/output flows and embedding spaces. 

\subsection{Hybrid hyperbolic-Euclidean models}\label{sec:hybrid_hypLLM}
\begin{figure}[ht]
    \centering
    \vspace{-.14in}
    \includegraphics[width=.79\textwidth]{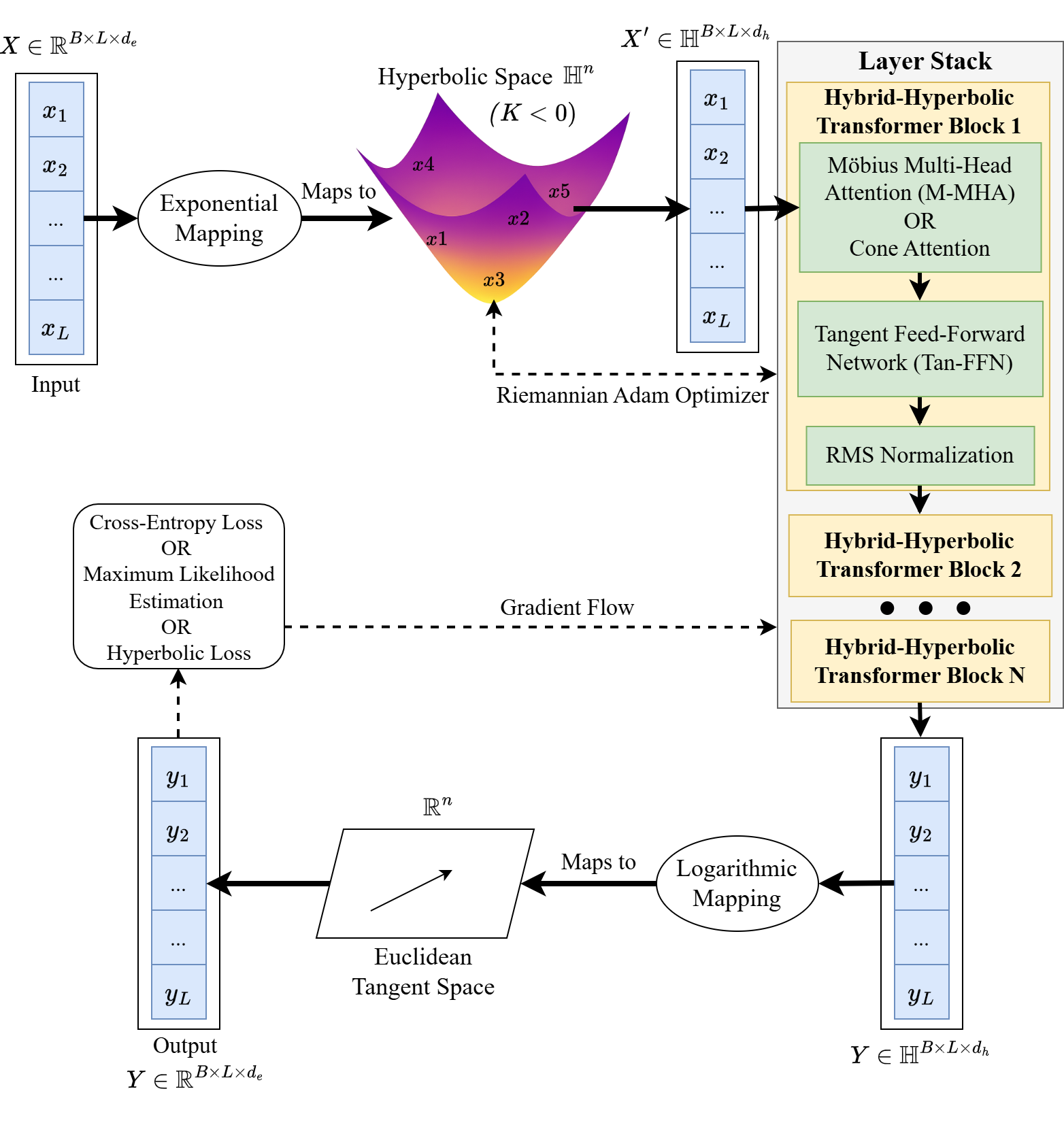}
    \caption{Hybrid models use exponential/logarithmic mappings between Euclidean and hyperbolic spaces with options for Poincar\'e ball or Lorentz embeddings.}
    \label{fig:hybrid_architecture}
    \vspace{-.22in}
\end{figure}

As illustrated in \Cref{fig:hybrid_architecture}, hybrid hyperbolic-Euclidean models represent the most common approach to integrating hyperbolic geometry into large language models. These architectures maintain standard Euclidean operations in most components while strategically incorporating hyperbolic representations in specific layers or modules through exponential/logarithmic mappings. The typical workflow involves: (1) mapping Euclidean input embeddings $X \in \mathbb{R}^{B \times L \times d_e}$ to hyperbolic space $\mathbb{H}^n$ via exponential maps (Poincar\'e ball or Lorentz lifting), (2) performing geometric operations in the curved space, and (3) mapping back to Euclidean space through logarithmic maps for compatibility with standard transformer components.

Hyperbolic BERT~\cite{chen2024hyperbolic} models standard BERT entirely within hyperbolic space by mapping dependency trees with geodesic distances on the Lorentz model. 
While Hyperbolic BERT demonstrates superior performance on tasks requiring hierarchical reasoning, it operates approximately 1.3 times slower than standard BERT due to the added computational overhead of manifold operations; additionally, it can exhibit numerical instabilities during repeated transformations. 
In this model, attention computations leverage hyperbolic distances, where each transformer layer maintains this hybrid hyperbolic-Euclidean structure, with feed-forward networks operating in tangent spaces to preserve computational efficiency while capturing hierarchical relationships through curved geometry. This computational overhead highlights the delicate balance between geometric expressiveness and practical deployability in hyperbolic LLMs. Another notable effort is HiT~\cite{he2024language}, which extends Transformer-based LLMs to encode linguistic hierarchies in a Poincar\'e ball. HiT similarly embeds tokens in a hyperbolic latent space to capture parse-tree structures, reporting improved entailment and transitive reasoning. 

In a multimodal context, a related approach, PoinCLIP~\cite{Srivastava2024} projects CLIP’s image-text embeddings into a Poincar\'e ball to enforce hierarchical distances, yielding improved zero-shot classification on coarse-grained categories. 
In summary, the exp/log map techniques in these models integrate hyperbolic geometry within the model’s forward pass (at multiple layers), rather than only at the input or output embedding layer, effectively making the network partially hyperbolic. Having established the mathematical foundations for hybrid hyperbolic-Euclidean models, we now also examine parameter-efficient approaches that adapt pre-trained LLMs to hyperbolic space through targeted fine-tuning strategies.


\subsection{Hyperbolic fine-tuned models}\label{sec:finetune_hypLLM}

\begin{figure}[ht]
    \centering
    \vspace{-.12in}
    \includegraphics[width=.79\textwidth]{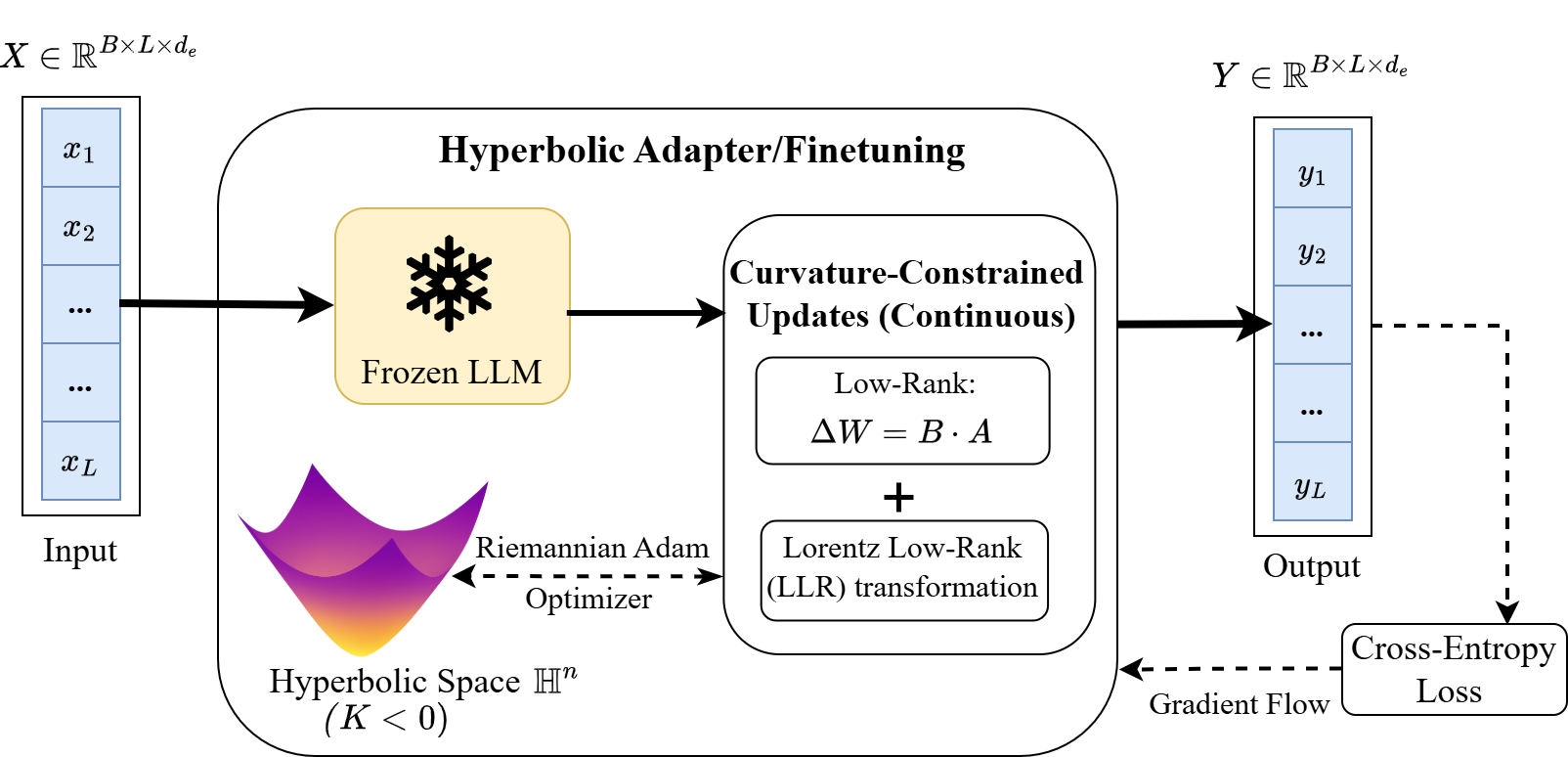}
    \caption{Fine-tuned approaches adapt frozen pre-trained models with hyperbolic adapters using curvature-constrained updates.}
    \label{fig:finetuning_architecture}
    \vspace{-.2in}
\end{figure}

Recent research has explored methods for fine-tuning pre-trained LLMs in hyperbolic space to better capture hierarchical structures without retraining entire models from scratch. Hyperbolic fine-tuned models represent a special type of hybrid hyperbolic-Euclidean models (section.~\ref{sec:hybrid_hypLLM}) that adapt frozen pre-trained architectures to hyperbolic space rather than integrating hyperbolic operations within the forward pass. As depicted in \Cref{fig:finetuning_architecture}, hyperbolic fine-tuned models represent a parameter-efficient approach to adapting pre-trained LLMs for hyperbolic geometry. The architecture centers around a frozen LLM backbone with specialized hyperbolic adapters that perform curvature-constrained updates. 

Token embeddings from a pre-trained GPT model have been observed~\cite{yang2024hyperbolic} to exhibit a high degree of inherent hyperbolicity with frequent tokens clustering near the origin and rare tokens farther out, indicating a latent tree structure. To exploit this, HypLoRA proposes a parameter-efficient fine-tuning method that avoids cancellation effects from naive exponential/logarithmic map applications. HypLoRA performs low-rank adaptation on the hyperbolic manifold without utilizing tangent space. Given input $\mathbf{x}_E \in \mathbb{R}^{d_e}$ and a frozen weight 
matrix $W$, HypLoRA introduces low-rank adaptation matrices $B \in \mathbb{R}^{d_e \times r}$ and $A \in \mathbb{R}^{r \times d_e}$ with rank $r \ll d_e$ that approximate the weight update $\Delta W \approx BA$. 
The Lorentz Low-Rank (LLR) transformation is:
\vspace{-0.15cm}
\begin{equation}
\mathbf{z}_E = W\mathbf{x}_E + \log_0\Big(\text{LLR}\big(BA, \exp_0(\mathbf{x}_E)\big)\Big),
\label{eq:hyplora}
\end{equation}

where $\mathbf{z}_E$ denotes the intermediate output in Euclidean space, $\exp_0(\cdot)$ and $\log_0(\cdot)$ are the exponential and logarithmic maps at the origin $\mathbf{o}$, and $\text{LLR}(BA, \mathbf{x}_H)$ denotes the direct Lorentz Low-Rank transformation on hyperbolic representation $\mathbf{x}_H$, while the pre-trained weights $W$ remain frozen. The final model output $Y$ is obtained after passing through multiple such layers in the frozen LLM backbone. Unlike the naive LoRA approach, which suffers from cancellation where the $\Delta W$ term vanishes due to curvature distortion, HypLoRA's manifold-based operations preserve the hyperbolic structure. HypLoRA achieves 13\% improvement on the AQuA mathematical reasoning benchmark~\cite{yang2024hyperbolic}, though its fundamental limitation is that the base model remains Euclidean, with hyperbolic operations confined to adapter modules.

Building upon the HypLoRA, HoRA (Hyperbolic Low-Rank Adaptation) was developed~\cite{yang2024enhancing} using the same direct Lorentz Low-Rank (LLR) formulation but with \emph{adaptive curvature}. It achieves 17.30\% improvement over Euclidean LoRA on mathematical reasoning tasks, the HoRA's hyperbolic component is essentially a wrapper around Euclidean weights. Notably, these hyperbolic fine-tuning methods have demonstrated significant gains on complex reasoning tasks (e.g., math word problem solving). Despite curvature-scaled training, the ``hyperbolic vanishing gradient'' problem can persist due to residual Euclidean computations during backpropagation. Nonetheless, the success of these approaches shows that pre-trained models encode latent hierarchical structures that hyperbolic fine-tuning can exploit. Although fine-tuning is computationally efficient, it still relies on Euclidean operations. To fully leverage hyperbolic geometry's representational power, we next explore architectures that operate entirely within the hyperbolic space.

\subsection{Fully hyperbolic models}
\begin{figure}[ht]
    \centering
    \vspace{-.1in}
    \includegraphics[width=.88\textwidth]{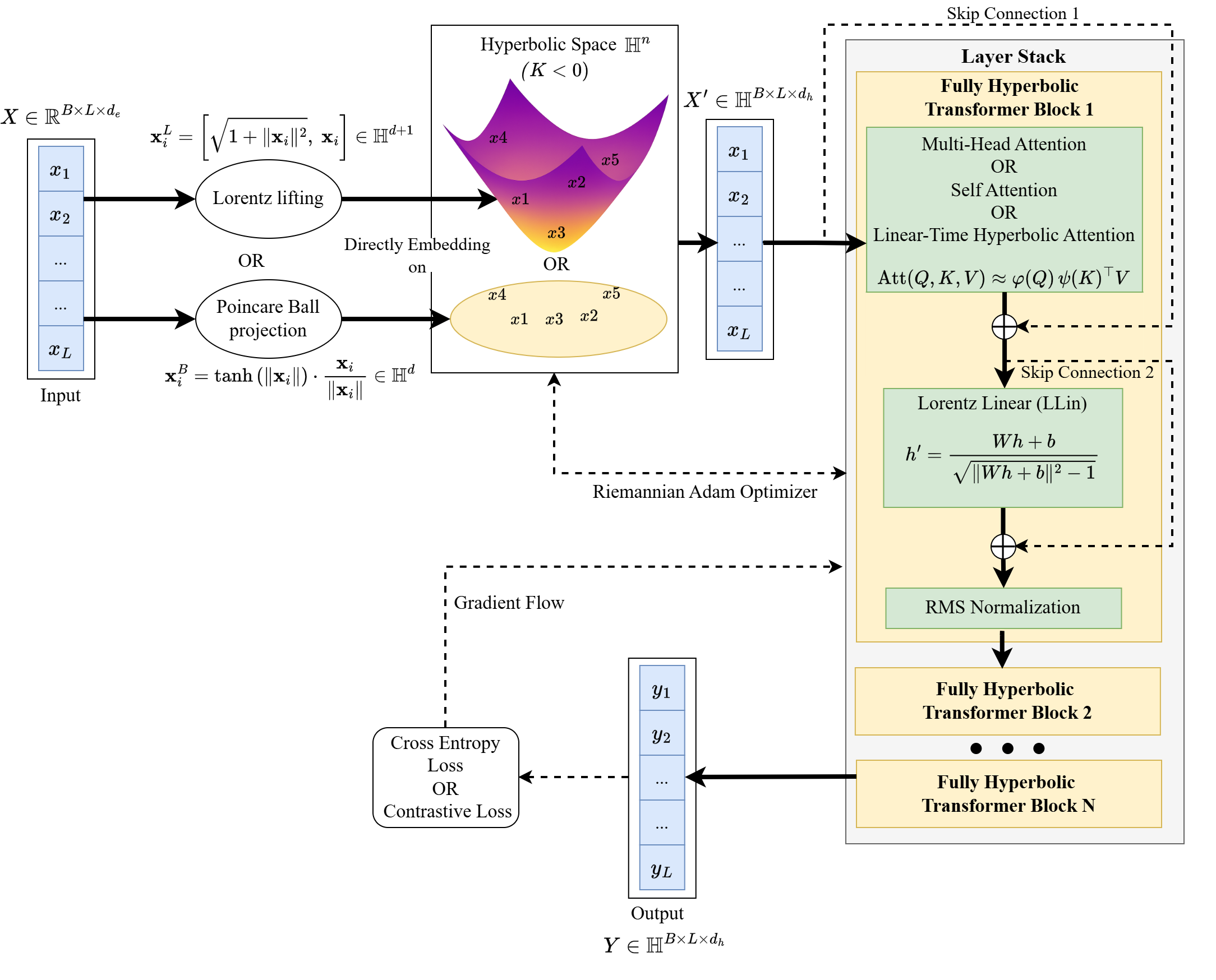}
    \caption{Fully hyperbolic models operate entirely in curved space with specialized geometric operations including linear-time hyperbolic attention and Lorentz linear transformations.}
    \label{fig:fully_hyperbolic_architecture}
    \vspace{-.2in}
\end{figure}

Unlike hybrid models that rely on exponential/logarithmic mappings, these architectures operate entirely within hyperbolic space $\mathbb{H}^n$ using specialized geometric operations. The architecture of fully hyperbolic models illustrated in \Cref{fig:fully_hyperbolic_architecture} represents the most theoretically complete approach to hyperbolic LLMs.  The workflow shows how the Euclidean input embeddings $X \in \mathbb{R}^{B \times L \times d_e}$ are processed through fully hyperbolic transformer blocks containing linear-time hyperbolic attention, Möbius multi-head attention or cone attention mechanisms, and Lorentz linear transformations, ultimately producing hyperbolic outputs $Y \in \mathbb{H}^{B \times L \times d_h}$.

Notably, Hypformer was proposed~\cite{yang2024hypformer} using specialized hyperbolic transformation blocks (HTC and HRC) that operate directly on the Lorentz manifold, avoiding frequent exponential/logarithmic mappings between tangent and hyperbolic spaces. Hypformer achieves linear complexity through hyperbolic factorization of the attention mechanism. The traditional quadratic attention $\mathcal{O}(N^2)$ is reduced to linear $\mathcal{O}(N)$ by changing the computation order from $(Q^TK)V$ to $Q(K^TV)$, where $Q$ is the query, $K$ is the key, and $V$ is the value:
$
\text{Attention}(Q,K,V) \approx \phi(Q) \psi(K)^T V,
\label{eq:hypformer_linear}
$ where $\phi(\cdot)$ and $\psi(\cdot)$ are hyperbolic feature maps that preserve the Lorentzian structure while enabling efficient computation through kernel approximation methods. 
It's linear-time ``hyperbolic attention'' mechanism eliminates the computational bottleneck of traditional quadratic attention while preserving the geometric properties essential for hierarchical modeling. Hypformer achieves superior results over both Euclidean and hybrid hyperbolic-Euclidean transformers on hierarchical reasoning tasks. One of the recent works introduced \textit{HELM}~\cite{he2025helm}, a Hyperbolic LLM that operates ``fully in hyperbolic space'' while employing a Mixture-of-Curvature Experts design. Instead of using a single constant curvature for the entire model, HELM assigns different transformer experts to different curvature spaces (including hyperbolic subspaces with varying radii), allowing the model to adaptively represent features at multiple scales of hierarchy. 
Early results show that HELM achieves competitive or better accuracy than Hypformer on challenging reasoning tasks, with the added ability to model diverse structural patterns. 
This architectural flexibility addresses a key limitation of fixed-curvature hyperbolic models. Furthermore, fully hyperbolic frameworks based on Lorentz transformations have been proposed~\cite{chen2021fully} to overcome the limitations of hybrid architectures that rely on tangent space operations. These approaches perform all operations directly in hyperbolic space, eliminating the need for exponential and logarithmic maps:
\begin{equation}
\mathbf{h}' = \text{LorentzLinear}(\mathbf{h}) = \frac{(W\mathbf{h} + b)}{\sqrt{\|W\mathbf{h} + b\|^2 - 1}}.
\end{equation}

This formula rescales the usual linear combination $W\mathbf{h}+b$ onto the hyperboloid (ensuring it lies in $\{x: \langle x,x\rangle_L=-1\}$). 
The Lorentz-Linear transformation maintains the hyperbolic constraint through normalization, ensuring that all intermediate representations remain on the hyperboloid manifold throughout the forward pass. 
More recently, \textit{HyperCore} was introduced~\cite{he2025hypercore}, an open-source framework providing a comprehensive suite of {\it fully hyperbolic neural modules}. These generalize standard deep learning layers (MLPs, CNNs, GNNs, and Transformers) to hyperbolic space, supporting both Lorentz and Poincar\'e manifolds. A key innovation is that every operation-attention, normalization, dropout, and feed-forward layers-remains within the hyperbolic manifold, avoiding tangent space projections. Specifically, HyperCore implements Lorentzian variants of core Transformer components, such as \texttt{LTransformerBlock}, which uses \textit{Lorentz multi-head attention} and \textit{Lorentz linear layers}. 
The success of fully hyperbolic transformers has inspired researchers to explore alternative architectures. State-space models, with their linear scalability, present unique opportunities for hyperbolic adaptation while maintaining computational efficiency.

\subsection{Hyperbolic state-space models}\label{sec:ssm_hypLLM}

Beyond Transformer-based architectures, fully hyperbolic sequence models have also been explored using state-space architectures. These models address the quadratic complexity limitations of hyperbolic transformers while maintaining the ability to capture hierarchical relationships across long sequences. For instance, Hierarchical Mamba (HiM)~\cite{patil2025hierarchical}, integrates the linear-time Mamba2 sequence model with hyperbolic geometry. HiM projects sequence representations into a Poincar\'e ball or Lorentz manifold with a learnable curvature and enforces hierarchy through specialized hyperbolic loss functions, while explicitly constraining norms to ensure numerical stability. HiM introduces explicit hierarchical constraints via learned hyperbolic geometry. It uses a learnable norm-scaling factor $\gamma$ and learnable curvature $\mathcal{K}$ so that scaling the pooled embedding by $\gamma$ before projection effectively lets the model adaptively find the optimal curvature and scale for each hierarchy level. HiM also uses a dynamic centripetal loss and clustering loss: the centripetal loss forces parent-node embeddings to lie closer to the hyperbolic origin than their children, and the clustering loss groups related points while pushing unrelated ones apart. 

These dynamic hyperbolic losses update their centripetal and clustering margins based on the updated radius as the curvature $\mathcal{K}$ keeps adapting. In effect, HiM’s mathematical innovation is to jointly learn a hyperbolic embedding scale and curvature while preserving tree-like structure end-to-end and scales long sequences as per state-space duality, yielding significant improvements on hierarchical reasoning benchmarks (mixed-hop prediction, multi-hop inference), compared to equivalent Euclidean models. 

Recent advances in state-space models have led to the development of the HMamba~\cite{zhang2025hmamba}, which integrates hyperbolic geometry with Mamba's selective state-space mechanism for the sequential recommendation task. HMamba introduces two distinct architectural variants: HMamba-Full and HMamba-Half. HMamba-Full represents a fully hyperbolic architecture that processes both input and output entirely in hyperbolic space. In contrast, HMamba-Half is a partially hyperbolic variant that implements a hybrid approach using Euclidean space to balance computational efficiency with hyperbolic benefits.
HMamba-Full implements curvature-normalized discretization with learnable curvature $\mathcal{K} < 0$, where hidden states evolve via Lorentz parallel transport to maintain hierarchical geometry throughout temporal evolution.
By contrast, HMamba-Half scores sequences in flat Euclidean space, so it does not use any curvature-dependent operation or PTrans. This hybrid design demonstrates the trade-off between geometric expressiveness and computational efficiency: HMamba-Full achieves superior hierarchical modeling at the cost of increased computational overhead, while HMamba-Half provides a balance suitable for large-scale applications where full hyperbolic computation may be prohibitive. \textit{SHMamba} (Structured Hyperbolic State Space Model) was introduced~\cite{yang2024shmamba} as a novel approach that combines hyperbolic geometry with state space models for audio-visual question answering (AVQA). 
Experimental results demonstrate that SHMamba achieves superior performance with 78.12\% fewer parameters while improving average performance by 2.53\% compared to transformer-based baselines, validating the effectiveness of combining hyperbolic geometry with state space architectures for multimodal understanding tasks~\cite{yang2024shmamba}. By operating entirely in hyperbolic space, these models better preserve hierarchical relationships throughout the network, leading to improved performance on tasks that require modeling complex hierarchical structures. 

We also provide a general comparison of the computational trade-offs and design choices across these categories in \cref{tab:hypllm_category_comparison_compact}. Having established the architectural foundations of hyperbolic LLMs, we now examine how these models perform with benchmarks along with their challenges and existing solutions.

\vspace{-0.5cm}

\begin{table*}[!h]
\centering
\setlength{\tabcolsep}{3pt}  
\renewcommand{\arraystretch}{1.0}
\caption{Comparative summary of advantages, limitations, and computational trade-offs among hybrid (Hyp–Eucl) models, hyperbolic fine-tuned models, fully hyperbolic models, and state-space models. Symbols: $\checkmark$ = supported, $\times$ = not supported, $\circ$ = partial or varying behavior.}
\label{tab:hypllm_category_comparison_compact}
{\footnotesize
\begin{tabular}{|l|c|c|c|c|}
\hline
\textbf{Property} & 
\textbf{\makecell{Hybrid \\Hyp–Eucl}} & 
\textbf{\makecell{Hyperbolic \\Fine-tuned}} & 
\textbf{\makecell{Fully \\Hyperbolic}} & 
\textbf{\makecell{Hyperbolic \\SSM}} \\
\hline
Linear time complexity $\mathcal{O}(N)$                 & $\times$     & $\times$     & $\circ$     & $\checkmark$ \\
\hline
Uses exp/log maps                                        & $\checkmark$ & $\checkmark$ & $\circ$     & $\circ$      \\
\hline
Fully hyperbolic                                         & $\times$     & $\times$     & $\checkmark$& $\circ$      \\
\hline
Frozen pre-trained backbone                              & $\times$     & $\checkmark$ & $\times$    & $\circ$      \\
\hline
Long-sequence modeling                                   & $\times$     & $\times$     & $\circ$     & $\checkmark$ \\
\hline
Hierarchical reasoning                                   & $\checkmark$ & $\checkmark$ & $\checkmark$& $\checkmark$ \\
\hline
\end{tabular}
}
\end{table*}

\section{Benchmarking and evaluation of hyperbolic LLMs}
\label{sec:benchmarking}

The evaluation of hyperbolic large language models presents unique challenges due to their specialized geometric properties and diverse architecture integrations of hyperbolic geometry. This section examines performance benchmarks of hyperbolic LLMs on mathematical reasoning and hierarchical language reasoning tasks, highlighting current benchmarking limitations and the need for a unified evaluation framework.

\subsection{Performance benchmarks across HypLLMs}

\subsubsection*{(1) Mathematical reasoning and arithmetic tasks}

Pre-trained language models exhibit strong intrinsic hyperbolic structure, as quantified in \cref{tab:hyperbolicity_analysis} through $\delta$-hyperbolicity measurements across architectures and datasets represent values approaching zero. Dataset's tree-likeness can be quantified via Gromov $\delta$-hyperbolicity; details and computation appear in Appendix~\ref{sec:gromov-delta}. Leveraging this intrinsic tree-like geometry, Hyperbolic fine-tuned LLMs like HypLoRA (as detailed in Section~\ref{sec:finetune_hypLLM}) operate through tangent space projections, enabling parameter-efficient adaptation that leverages this intrinsic tree-like geometry. HypLoRA have demonstrated exceptional performance on complex reasoning tasks, e.g., the results in \cref{tab:math_reasoning_results} demonstrate that HypLoRA~\cite{yang2024hyperbolic} consistently outperforms both standard LoRA~\cite{hu2022lora} and DoRA~\cite{liu2024dora} across multiple mathematical reasoning datasets, with particularly significant improvements on the most challenging AQuA dataset (13.0\% improvement for Gemma-7B), validating the effectiveness for mathematical reasoning.


\begin{table}[ht]
\centering
\caption{$\delta$-Hyperbolicity analysis across LLM architectures and datasets~\cite{yang2024hyperbolic}.}
\label{tab:hyperbolicity_analysis}
\setlength{\tabcolsep}{4pt}
\footnotesize
\begin{tabular}{|l|c|c|c|c|}
\hline
\textbf{Backbone} & \textbf{MAWPS~\cite{koncel2016mawps}} & \textbf{SVAMP~\cite{patel2021nlp}} & \textbf{GSM8K~\cite{cobbe2021training}} & \textbf{AQuA~\cite{ling2017program}} \\
\hline
LLaMA-7B & $0.08 \pm 0.02$ & $0.09 \pm 0.01$ & $0.10 \pm 0.01$ & $0.10 \pm 0.01$ \\
LLaMA-13B & $0.08 \pm 0.01$ & $0.09 \pm 0.01$ & $0.09 \pm 0.01$ & $0.10 \pm 0.01$ \\
Gemma-7B & $0.11 \pm 0.01$ & $0.11 \pm 0.01$ & $0.11 \pm 0.01$ & $0.12 \pm 0.01$ \\
LLaMA3-8B & $0.06 \pm 0.01$ & $0.07 \pm 0.01$ & $0.07 \pm 0.01$ & $0.08 \pm 0.01$ \\
\hline
\textbf{Average $\delta$} & $0.08 \pm 0.01$ & $0.09 \pm 0.01$ & $0.09 \pm 0.01$ & $0.10 \pm 0.01$ \\
\hline
\end{tabular}
\vspace{-.10in}
\end{table}

\begin{table}[h]
\vspace{-.2in}
\centering
\caption{Mathematical reasoning performance: HypLoRA vs. baseline methods (Accuracy \%) ~\cite{yang2024hyperbolic}.}
\label{tab:math_reasoning_results}
\setlength{\tabcolsep}{3pt}
\footnotesize
\begin{tabular}{|l|c|c|c|c|c|}
\hline
\textbf{Backbone} & \textbf{Model} & \textbf{MAWPS~\cite{koncel2016mawps}} & \textbf{SVAMP~\cite{patel2021nlp}} & \textbf{GSM8K~\cite{cobbe2021training}} & \textbf{AQuA~\cite{ling2017program}} \\
\hline
\multirow{3}{*}{LLaMA-7B}
& LoRA & 79.0 & 52.1 & 37.5 & 18.9 \\
& DoRA & \textbf{80.0} & 48.8 & 39.0 & 16.4 \\
& \textbf{HypLoRA} & 79.0 & \textbf{49.1} & \textbf{39.1} & \textbf{20.5} \\
\hline
\multirow{3}{*}{LLaMA-13B}
& LoRA & \textbf{83.6} & 54.6 & 47.5 & 18.5 \\
& DoRA & 83.0 & 54.6 & --- & 18.9 \\
& \textbf{HypLoRA} & 83.2 & \textbf{54.8} & \textbf{49.0} & \textbf{21.5} \\
\hline
\multirow{3}{*}{Gemma-7B}
& LoRA & 91.6 & 76.2 & 66.3 & 28.9 \\
& DoRA & \textbf{91.7} & 75.9 & 65.4 & 27.7 \\
& \textbf{HypLoRA} & 91.5 & \textbf{78.7} & \textbf{69.5} & \textbf{32.7} \\
\hline
\multirow{3}{*}{LLaMA3-8B}
& LoRA & \textbf{92.7} & 78.9 & 70.8 & 30.4 \\
& DoRA & 92.4 & 79.3 & 71.3 & 33.1 \\
& \textbf{HypLoRA} & 91.6 & \textbf{80.5} & \textbf{74.0} & \textbf{34.2} \\
\hline
\end{tabular}
\vspace{-.2in}
\end{table}

\subsubsection*{(2) Hierarchical language reasoning (mixed-hop prediction) task}

Hierarchical language reasoning tasks evaluate models' ability to understand and predict relationships across different levels of semantic hierarchies.  The \textbf{mixed-hop prediction task} is particularly challenging as it requires models to predict the exact hierarchical distance between entities in ontological structures, distinguishing between direct (1-hop) and indirect multi-hop relationships. In this subsection, we evaluate two representative models from distinct architectural categories established in Section 3: \textbf{Hierarchical Transformer (HiT)}~\cite{he2024language}, a hybrid hyperbolic-Euclidean model (Section~\ref{sec:hybrid_hypLLM}), and \textbf{Hierarchical Mamba (HiM)}~\cite{patil2025hierarchical}, a hyperbolic state-space model (Section~\ref{sec:ssm_hypLLM}) that operates entirely in curved space with learnable curvature. \textbf{Hierarchical Mamba (HiM)}~\cite{patil2025hierarchical} demonstrates exceptional performance in this domain by integrating efficient Mamba2 state-space models with hyperbolic geometry.
Our experimental evaluation results for mixed-hop prediction tasks across ontological datasets are presented in \Cref{tab:hierarchical_language_results} that compares hyperbolic models against the Euclidean baseline \textit{SentenceMamba-16M}~\cite{patil2025hierarchical}. SentenceMamba-16M is a Mamba2-based large language model with 16 million parameters designed to generate high-quality sentence embeddings. Notably, HiM-Poincar\'e achieves the best performance across all datasets, highlighting the effectiveness of hyperbolic geometry combined with state space models for capturing hierarchical language structures. It should be noted that the original HiT uses a pretrained model (e.g., all-MiniLM-L6-v2) for hyperbolic space training with curvature $\mathcal{K} =-1/384$, which is slightly Euclidean-like. In our experiments, we use HiT-Rand-Init with random initialization and higher curvature $\mathcal{K}=-1.0$ for more hyperbolic behavior. We use this randomly initialized variant of HiT for fair comparison with HiM, since HiM is also trained with random initialization.

\begin{table}[ht]
\vspace{-.12in}
\centering
\caption{Mixed-hop prediction performance (F1 score \%): Comparison of Hierarchical Mamba (HiM) and Hierarchical Transformer (HiT) with Euclidean baseline SentenceMamba-16M across ontological datasets.}
\label{tab:hierarchical_language_results}
\setlength{\tabcolsep}{4pt}
\footnotesize
\begin{tabular}{|l|c|c|c|c|}
\hline
\textbf{Backbone} & \textbf{Model} & \textbf{WordNet} & \textbf{DOID} \\
\hline
SentenceMamba-16M (Mamba-2 based) & Euclidean baseline & 61.5 & 43.6\\
all-MiniLM-L6-v2 (Transformer-based) & HiT-Rand-Init Poincar\'e & 84.6 & 83.7 \\
SentenceMamba-16M (Mamba-2 based) & \textbf{HiM-Poincar\'e} & \textbf{85.9} & \textbf{90.2} \\
\hline
\end{tabular}
\vspace{-.12in}
\end{table}







\subsection{Benchmarking limitations}

Existing benchmarking approaches reveal several limitations that need to be addressed in future research: 

\textbf{(1) Hierarchical structure evaluation:} Existing benchmarks primarily focus on task-specific accuracy rather than measuring the quality of hierarchical representations. Future benchmarks should incorporate metrics that explicitly evaluate hierarchical structure preservation and multi-scale reasoning capabilities.

\textbf{(2) Scalability benchmarks:} Limited evaluation of hyperbolic models on large-scale datasets restricts understanding of scalability characteristics. Future benchmarking should include systematic evaluation across varying dataset sizes and model scales.

\textbf{(3) Unified evaluation framework:} The absence of standardized evaluation protocols across HypLLM categories complicates comparative analysis. Development of unified benchmarking frameworks would enable more robust cross-model comparisons and accelerate research progress.

The empirical evidence consistently demonstrates that hyperbolic geometry provides significant advantages for tasks involving hierarchical structures, complex reasoning, and multi-modal understanding, while revealing important areas for continued research and development in benchmarking methodologies.

\section{Main challenges in building hyperbolic LLMs}
\label{sec:challenges}

Developing large language models in hyperbolic space requires addressing fundamental challenges that stem from the incompatibility between curved manifolds and standard neural network operations. A major challenge in developing large language models (LLMs) in hyperbolic space lies in generalizing Euclidean operations to hyperbolic manifolds. Unlike simple element-wise operations (e.g., addition, multiplication) or fundamental neural network components such as linear transformations, attention mechanisms, and normalization, these operations and components require careful geometric reformulation in hyperbolic space~\cite{ganea2018hyperbolic,chen2021fully}. Even operations like convolution and pooling must be redesigned to respect the curved geometry of hyperbolic space, as standard Euclidean formulations do not preserve the geometric properties essential for hierarchical representation~\cite{yang2024hypformer}. The fundamental issue stems from the fact that hyperbolic manifolds do not naturally support the linear algebraic operations that underpin modern neural network architectures. This necessitates the development of specialized geometric operations and optimization techniques that respect the curved geometry of hyperbolic space while maintaining computational efficiency. Despite the theoretical advantages of hyperbolic geometry for modeling hierarchical data, developing practical hyperbolic large language models faces several fundamental challenges that significantly impact their scalability and deployment. Table~\ref{tab:challenges_summary} summarizes these challenges, their technical manifestations, current solutions, and promising future directions.

\begin{table}[ht]
\centering
\caption{Summary of main challenges and future research directions in hyperbolic LLMs.}
\label{tab:challenges_summary}
\small
\setlength{\tabcolsep}{4pt}
\begin{tabular}{p{2.1cm}p{4.8cm}p{4.8cm}}
\toprule
\textbf{Type} & \textbf{Challenges} & \textbf{Potential Solutions} \\
\midrule
\makecell[tl]{Numerical \\Stability} & 
\makecell[tl]{• Poincar\'e: $r_0 \approx 38$ (64-bit), \\
$\mathcal{O}(\epsilon^2)$ vanishing\\
• Lorentz: $r_0 \approx 2^{19}$ limit \\
} &
\makecell[tl]{• Gradient rescaling \\
• Weighted lorentz centroids \\
• Curvature-sensitive\\ initialization \\
• Euclidean parametrization \\
• Unified stable parametrization}
\\[0.5em]
\midrule
Computational Overhead &
\makecell[tl]{• Exp/log maps: $\sim$30\% slower \\
• Attention: $\mathcal{O}(N^2)$ complexity \\
• Parallel transport costly \\
} &
\makecell[tl]{• Sparse attention \\
• Sparse spectral training \\
• Weighted lorentz centroids \\
• State-space alternatives}
\\[0.5em]
\midrule
Curvature Adaption &
\makecell[tl]{• Fixed curvature suboptimal for \\
varying depths \\
• Minimal updates during\\ training \\
} &
\makecell[tl]{• Learnable curvature \\
• Mixture-of-curvature experts \\
• Gradient-based optimization \\}
\\[0.5em]
\midrule
\makecell[tl]{Operation \\Design} &
\makecell[tl]{• No hyperbolic equivalents\\ for 
batch norm, dropout, and \\concatenation\\
• Manifold constraint violations \\
} &
\makecell[tl]{• Automated operation discovery \\
• Projection-free architectures \\
• Riemannian optimization\\ schemes}
\\[0.5em]
\midrule
\makecell[tl]{Hardware \\Support} &
\makecell[tl]{• Multi-component floats not \\
GPU-supported\\
• No framework native support \\
} &
\makecell[tl]{• Multi-precision floats \\
• Memory-efficient\\ representations \\
• Quantization techniques \\
• Hardware-aware geometric \\operations \\}
\\[0.5em]
\bottomrule
\end{tabular}
\end{table}

\textbf{Numerical Stability} 
One of the most critical challenges in hyperbolic LLMs stems from the inherent numerical instabilities of hyperbolic representations in floating-point arithmetic. The exponential volume growth property of hyperbolic space leads to catastrophic numerical problems. In the Poincar\'e ball, points can only be accurately represented within a ball of radius $r_0 \approx 38$ under 64-bit floating-point precision, as boundary proximity causes gradient vanishing with magnitude scaling as $\mathcal{O}(\epsilon^2)$ for points at distance $\epsilon$ from the boundary~\cite{pmlr-v202-mishne23a}. The Lorentz model's closed-form geodesic computation partially mitigates these issues but still faces representation capacity limitations within radius $r_0 \approx 2^{19}$. Current solutions include Euclidean parametrization~\cite{pmlr-v202-mishne23a} and careful initialization strategies~\cite{yu2021representing}, though multi-precision arithmetic and unified stable parametrizations remain active research directions.

\textbf{Computational Overhead} 
Hyperbolic LLMs face significant computational challenges that limit practical deployment. The frequent use of exponential and logarithmic mappings introduces substantial overhead, with training times approximately 30\% slower than standard Euclidean models~\cite{chen2024hyperbolic,yang2024hyperbolic}. Traditional hyperbolic attention mechanisms suffer from quadratic complexity $\mathcal{O}(N^2)$, becoming particularly problematic for long-sequence modeling. Recent advances include linear-time hyperbolic attention through sparse attention, sparse spectral training or state-space models, though hardware-accelerated geometric operations and hierarchical attention patterns present promising future directions. Recent work on Lorentzian residual networks~\cite{he2025lorentzian} addresses these computational challenges by eliminating parallel transport operations entirely, operating directly on the hyperbolic manifold through weighted Lorentzian centroids. This approach achieves over 2000$\times$ speedup compared to parallel transport methods while ensuring numerical stability through normalized projections, demonstrating that efficient hyperbolic operations are achievable without sacrificing geometric properties.

\textbf{Curvature Adaption} 
Choosing appropriate curvature values for different layers and components remains a significant challenge. Fixed curvature models may not optimally represent data with varying hierarchical depths, while learnable curvature approaches introduce additional hyperparameters and optimization complexity~\cite{he2025helm}. Recent mixture-of-curvature approaches show promise but add substantial architectural complexity.

\textbf{Operation Design} 
Many standard neural network operations lack well-defined hyperbolic counterparts. Operations such as batch normalization, dropout, concatenation, and various activation functions require careful geometric reformulation~\cite{yang2024hypformer,chen2021fully}. Maintaining points on the hyperbolic manifold throughout training requires projection operations or constraint-preserving update schemes, introducing computational overhead and potential error accumulation. Recent frameworks like HyperCore~\cite{he2025hypercore} provide comprehensive operation libraries, though automated operation discovery and projection-free architectures represent important future directions.

\textbf{Hardware Support.} 
Hyperbolic operations often require higher precision arithmetic, leading to increased memory consumption. Multi-component floating-point representations~\cite{yu2021representing} are not supported on ML accelerators, forcing slow CPU computations. Current deep learning frameworks lack native support for hyperbolic operations, requiring custom implementations. Future work must address memory-efficient representations, quantization techniques, and dedicated hardware support to enable practical large-scale deployment.

These challenges collectively represent significant barriers to widespread adoption of hyperbolic LLMs. However, ongoing research continues to address these limitations through improved numerical methods, more efficient architectures, and better optimization techniques. Understanding these challenges is crucial for researchers seeking to develop robust and scalable hyperbolic language models that realize the theoretical benefits of hyperbolic geometry in practical applications.

\section{Emerging applications of HypLLMs}
\label{sec:applications}

\begin{figure*}[ht]
\centering
\includegraphics[width=\textwidth]{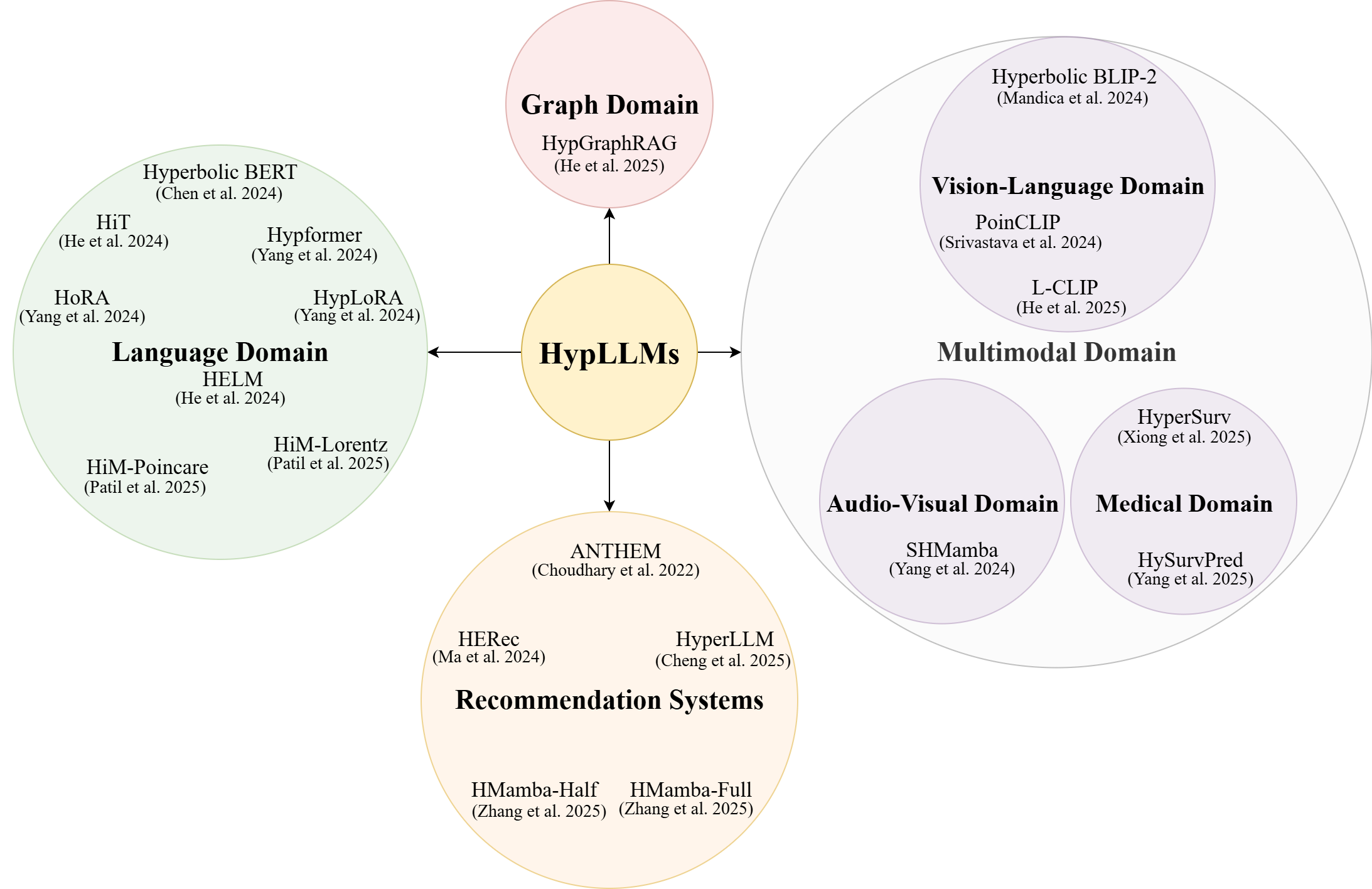}
    \caption{Applications of hyperbolic large language models (HypLLMs) across different domains. Each bubble corresponds to a domain, with included HypLLMs and their corresponding authors.}
    \label{fig:hypLLM_domains}
    \vspace{-.25in}
\end{figure*}

The theoretical foundations and architectural innovations in hyperbolic geometry have enabled HypLLMs to demonstrate remarkable versatility across diverse application domains. As illustrated in \Cref{fig:hypLLM_domains}, hyperbolic large language models have found successful applications spanning computer vision, natural language processing, multimodal representation learning, neuroscience, and biomedical domains. 
This section examines how HypLLMs have been successfully deployed across various domains, demonstrating both the practical utility and broad applicability of hyperbolic geometric principles in LLM architectures.

\subsection{HypLLMs for computer vision}
\label{sec:hllmscompvis}


Large-scale vision tasks often involve hierarchical relationships; for example, object recognition can be organized by a taxonomy of categories (from general classes to fine-grained subtypes). Hyperbolic representations have begun to show promise in computer vision by capturing these inherent hierarchies. One prominent approach is to incorporate hyperbolic geometry into vision-language models. \textit{PoinCLIP} was proposed~\cite{Srivastava2024} as a vision–language model that projects CLIP embeddings into the Poincar\'e ball to better reflect the conceptual hierarchy between images and text descriptions. By learning joint image-text representations entirely in a hyperbolic space, PoinCLIP preserves the natural “is-a” relationships (e.g., an image of a \emph{dog} is also an image of an \emph{animal}) as distances in the embedding space. Using hyperbolic image-text embeddings yields improved zero-shot classification and image retrieval performance over the standard (Euclidean) CLIP, especially for abstract or coarse-grained (hierarchical) categories. Building upon these foundational insights, more comprehensive hyperbolic vision-language architectures have emerged.
The BLIP-2 architecture has been extended~\cite{mandica2024hyperbolic} into hyperbolic space to capture hierarchical image-text relationships called \textit{Hyperbolic BLIP-2}. Their model uses a frozen image encoder and frozen LLM but projects visual and language embeddings into the Poincar\'e ball using the exponential map. 

Recent work~\cite{qu2024llms} proposes LLM-AR, a framework that projects skeleton sequences into a linguistic format for LLM-based action recognition, utilizing a VQ-VAE with a hyperbolic codebook for the encoder. While the LLM component in LLM-AR operates in Euclidean space, the use of hyperbolic geometry in the codebook highlights the growing convergence of hyperbolic representation learning and LLMs for structured vision tasks. In parallel, HyLiFormer~\cite{li2025hyliformer} demonstrates the advantages of fully learning in hyperbolic space for skeleton-based action recognition. Beyond vision-language alignment, hyperbolic embeddings have also been applied to purely visual tasks. For instance, image classifiers can be enhanced by replacing the final Euclidean feature layer with a hyperbolic space that encodes class hierarchies~\cite{dhall2020hierarchical}. This allows the model to naturally leverage hierarchical label structures, e.g., an image misclassified within the correct superclass is penalized less, leading to better generalization on imbalanced or hierarchical datasets.

\subsection{HypLLMs for sequence modeling}


Many sequential datasets have an intrinsic hierarchical organization, where hyperbolic geometry can be leveraged to model long-range dependencies or branching patterns. 
\textit{Poincar\'e maps} were introduced~\cite{klimovskaia2020poincare} to analyze single-cell gene expression sequences. In single-cell developmental trajectories, cells differentiate following a tree-like hierarchy with multiple branches. They showed that mapping high-dimensional scRNA-seq data into a 2D hyperbolic disk uncovers these branching hierarchies more faithfully than Euclidean PCA or UMAP, while preserving meaningful distances between cell states. Each \emph{Poincar\'e map} (hyperbolic embedding) effectively acts as a continuous analog of a lineage tree, enabling clearer visualization and more accurate clustering, pseudotime ordering, and lineage inference. This demonstrates the power of hyperbolic representations for sequence modeling in biology, where chronological or evolutionary sequences form implicit hierarchies. 

Recent work has begun applying hyperbolic LLM techniques to sequential data in NLP and beyond, using novel and efficient architectures. The HiM model~\cite{patil2025hierarchical} exemplifies how hyperbolic geometry can enhance language sequence modeling. By leveraging Mamba’s selective state-space mechanism, HiM projects outputs into a curved space (Poincar\'e or Lorentz) with learnable curvature and dynamic margins. Across various ontological datasets, HiM captures multi-level linguistic hierarchies more effectively than Euclidean baselines. Similarly, in the recommendation domain, \textit{Hyperbolic Mamba (HMamba)}\cite{zhang2025hmamba} is used to model user-item sequential interactions. HMamba unifies Mamba’s linear-time state-space architecture with hyperbolic representational power, implementing a curvature-aware state space and stabilized Riemannian operations to maintain hierarchy in user preference sequences. Thanks to this design, HMamba achieves 3–11\% higher top-$K$ recommendation accuracy than comparable Transformer-based methods, while retaining Mamba’s $\mathcal{O}(L)$ efficiency for long interaction streams. \textit{HERec}\cite{ma2024harec} is a hyperbolic graph-LLM framework that unifies semantic information from LLMs with collaborative user-item graphs in hyperbolic space
, achieving state-of-the-art performance on both utility and diversity metrics in recommendation tasks. These cases illustrate how HypLLM approaches can be extended to various sequence learning problems-from natural language to recommender systems-capturing latent hierarchical structures without sacrificing scalability.

A comparative analysis~\cite{pandey2024comparative} reports that for dynamic graph sequences (i.e., temporal networks), Mamba-based models can match or exceed Transformer performance in link prediction tasks while being significantly more efficient. Although their work does not incorporate hyperbolic geometry, it highlights the potential of state-space architectures in temporal graph modeling, suggesting that future work could explore integrating hyperbolic embeddings into such models to better capture evolving hierarchical structures in dynamic networks. In summary, HypLLM approaches for sequence data exploit hyperbolic space’s ability to naturally embed tree-like progression, whether it is cell differentiation lineages or abstract syntax trees in text, leading to richer and more interpretable sequence representations.

\subsection{HypLLMs for multimodal representation learning}


Hyperbolic LLM techniques are also being applied to multimodal representation learning problems, where data from different modalities (text, vision, graphs, etc.) have underlying hierarchical correspondences. A model called \textit{HyperSurv} has been proposed~\cite{xiongenhancing} that fuses pathology images and text reports for cancer survival prediction. In this task, doctors’ free-text reports describe visual findings in whole-slide images (WSIs); the descriptions have a hierarchical structure (e.g., broad terms like “tumor” entail specific attributes like “necrosis” or “mitosis count”). HyperSurv maps both the image features and the report embeddings into a shared hyperbolic space, using \emph{hyperbolic cones} to enforce entailment constraints. This approach naturally captures the one-to-many relationships between high-level medical terms and multiple image regions. As a result, the model achieved state-of-the-art accuracy in survival outcome prediction across several cancer datasets, outperforming Euclidean baselines by effectively modeling the multi-modal hierarchy of concepts and images. Building on similar principles for cancer survival prediction, HySurvPred was introduced~\cite{yang2025hysurvpred}, which integrates histopathology images and genomic data in hyperbolic space for survival prediction. Unlike HyperSurv's focus on pathology images and text reports, HySurvPred specifically addresses the hierarchical structures in both histopathology data (from tissue to cell scales) and genomics data (from biological networks to genes). 
The framework operates by mapping multimodal features from Euclidean space into hyperbolic space to better capture hierarchical relationships, achieving superior performance on five TCGA benchmark datasets and demonstrating the effectiveness of hyperbolic multimodal fusion for complex medical prediction tasks~\cite{yang2025hysurvpred}.

Another emerging direction is e-commerce and web multimodal data: Amazon researchers have explored hyperbolic embeddings for product catalogs, which include text descriptions, images, and user interaction graphs. By learning visuo-lingual concept embeddings in a hyperbolic space, a system can represent, say, a product image and its title within a shared hierarchy (where general product categories branch into specific products). \textit{ANTHEM} (Attentive Hyperbolic Entity Model) framework~\cite{choudhary2022anthem} was proposed, which models queries and products as hyperboloids in hyperbolic space and uses attention mechanisms to capture hierarchical and compositional relationships. Evaluated on large-scale e-commerce data, ANTHEM demonstrated more than 10\% improvement over state-of-the-art product search methods, highlighting the benefits of hyperbolic geometry for representing product knowledge graphs and “also-viewed” relationships with less distortion and in fewer dimensions. \textit{HyperLLM}~\cite{cheng2025large} is a model-agnostic framework that integrates large language models with hyperbolic space to capture hierarchical information in recommender systems. HyperLLM leverages LLMs to generate multi-level classification tags and aligns user-item interactions in hyperbolic space through contrastive learning, achieving over 40\% improvement compared to prior methods. Their work further validates HyperLLM’s effectiveness through detailed comparisons against multiple strong baselines. This work demonstrates the synergy between LLMs and hyperbolic geometry for hierarchical recommendation and training stability.

Similarly, the PoinCLIP model discussed in Section~\ref{sec:hllmscompvis} integrates hyperbolic geometry to better align images with text by their conceptual hierarchy, improving zero-shot recognition of image–text pairs \cite{Srivastava2024}. 
\textit{Hyperbolic BLIP-2} discussed earlier is one of the first billion-parameter hyperbolic multimodal LLMs~\cite{mandica2024hyperbolic}. In the audio-visual domain, SHMamba was proposed~\cite{yang2024shmamba}, demonstrating how hyperbolic geometry can enhance multimodal understanding in sequential data. The model addresses the inherent hierarchical structure present in audio-visual question-answering, where sounds and visual elements often follow tree-like organizational patterns (e.g., musical instrument categories branching into specific instruments, or object taxonomies in visual scenes). 
The adaptive curvature mechanism allows the model to flexibly explore different hierarchical structures depending on the specific audio-visual content characteristics, making it particularly effective for complex multimodal scenes with varying levels of semantic hierarchy. Evaluated on MUSIC-AVQA and AVQA datasets, SHMamba achieved state-of-the-art performance while maintaining computational efficiency, highlighting the potential of hyperbolic state space models for real-time multimodal applications. These examples illustrate that for multimodal problems involving inherently hierarchical data, hyperbolic LLMs provide a powerful tool. They enable a unified embedding space where different modalities meet, and where \textit{distance directly encodes semantic or conceptual similarity} across modalities, something especially beneficial when the modality interactions are complex and multi-scale.

\subsection{Hyperbolic embeddings for brain network analysis}
\label{sec:brain_networks}


Beyond language and knowledge graphs, hyperbolic embeddings have recently been applied in neuroscience to model the brain’s network organization. The concept of neural manifolds underlying brain function provides a crucial theoretical foundation for understanding why hyperbolic embeddings are particularly effective for neural data. Previous studies have demonstrated~\cite{gallego2017neural} that neural population activity during movement is constrained to low-dimensional manifolds spanned by specific patterns of correlated neural activity, termed ``neural modes.'' This manifold-based view of neural computation suggests that the coordinated activity of neural populations can be captured by relatively few covariation patterns, which naturally aligns with the hierarchical representational capacity of hyperbolic space. These manifolds have been shown~\cite{fortunato2024nonlinear} to exhibit intrinsically nonlinear structure, particularly during complex behaviors that require varied neural activity patterns. Their analysis across monkey, mouse, and human motor cortex revealed that nonlinear manifolds consistently outperform flat manifolds in capturing neural population dynamics, with the degree of nonlinearity varying across architecturally distinct brain regions. This finding has profound implications for hyperbolic neural models, as it suggests that the curved geometry of hyperbolic space may be fundamentally better suited to represent the intrinsic nonlinear structure of neural population activity than traditional Euclidean approaches.

A framework for embedding MEG (magnetoencephalography) functional brain networks into hyperbolic space has been developed~\cite{baker2024hyperbolic}, motivated by the observation that brain connectivity graphs possess a hierarchical structure (from local circuits to distributed modules) which might be captured more naturally in hyperbolic space $\mathbb{H}^n$. High-dimensional brain networks were successfully mapped into low-dimensional hyperbolic space while preserving both local and global geometric relationships with \textit{lower distortion} than competing Euclidean embedding methods. Even in 2 or 3 dimensions, the hyperbolic representation retained important network properties (connectivity patterns, community structure) that Euclidean PCA or spectral embeddings struggled to maintain. A new metric was defined in the latent space—the radial coordinate of a node's embedding-as a proxy for that node's hierarchical importance in the brain network~\cite{baker2024hyperbolic}. Using this metric, subtle but consistent differences were discovered between healthy individuals and those with subjective cognitive decline (SCD). In particular, participants with early cognitive decline showed a \textit{strong hierarchy} (greater average embedding radius) in specific subnetworks (e.g., dorsal and ventral attention networks) relative to healthy controls. These results suggest that the brains of SCD subjects exhibit a shift toward a more treelike, hierarchical organization, which hyperbolic embeddings can detect. Complementary advances in temporal modeling of brain connectivity using Mamba~\cite{huang2025brainatcl} demonstrate the potential of state-space models for efficient long-range dependency capture in dynamic functional connectivity analysis.

In addition to cognitive decline detection, hyperbolic neural models have recently been employed to model \textit{aging trajectories} in large-scale brain connectivity data. A Fully Hyperbolic Neural Network (FHNN) operating directly in Lorentz space was introduced to embed MEG-derived brain graphs from over 500 subjects~\cite{ramirez2024fully}. The model captures subtle, hierarchical reorganizations in subnetworks (e.g., attention, default mode networks) that occur with aging. This framework leverages the radius of Lorentz embeddings to quantify hierarchy and reveals that elderly individuals show a \textit{reduction} in embedding hierarchy-opposite to the increase observed in early cognitive decline. This application underscores the versatility of hyperbolic representations: they are not only mathematically appealing for theoretical data hierarchies, but also empirically powerful in analyzing real biological networks. By capturing multi-scale structure in brain connectivity data, hyperbolic LLM techniques can potentially contribute to cognitive modeling and clinical neuroscience. Such cross-disciplinary success strengthens the case that hyperbolic geometry is a unifying language for complex hierarchical systems across domains.

\section{Discussion}
The evolution of hyperbolic geometry from mathematical abstraction to practical computational tool represents a paradigm shift in representation learning for hierarchical data. Our taxonomy reveals distinct trade-offs across the four HypLLM categories: (1) hybrid hyperbolic-Euclidean models strategically integrate curved geometry into existing architectures through exponential/logarithmic mappings, proving effective for knowledge graph completion and taxonomy reasoning; (2) hyperbolic fine-tuned models provide parameter-efficient adaptation of pre-trained models, particularly successful in boosting logical reasoning and arithmetic problem-solving; (3) fully hyperbolic models operate entirely in curved space with specialized geometric operations, aimed at improving general NLP benchmark performance while maintaining computational efficiency; (4) hyperbolic state-space models combine the linear complexity of state-space architectures with hierarchical modeling capabilities, enabling efficient processing of long sequences with complex hierarchical structures.

Beyond traditional NLP and computer vision applications, recent neuroscience research has demonstrated that hyperbolic geometry provides fundamental insights into human cognitive processes and perception. Recent advances in Bayesian hyperbolic embedding methods~\cite{praturu2024adaptive} have enabled more principled approaches to modeling complex hierarchical data structures found in neural systems. A particularly promising direction lies in leveraging insights from low-dimensional, intrinsically nonlinear manifolds that provide more natural representation space for neural data~\cite{gallego2017neural,fortunato2024nonlinear}. A significant work~\cite{lee2024navigating} also revealed that object concept memorability follows hierarchical structures better captured by hyperbolic rather than Euclidean geometry, showing that memorable objects cluster near the center of hyperbolic representational spaces while forgettable objects scatter toward the periphery. This finding suggests that human memory systems may inherently operate within curved geometric frameworks that preserve hierarchical relationships among concepts. It has been demonstrated that Human olfactory perception can be accurately modeled using three-dimensional hyperbolic spaces, where natural odors and perceptual descriptions both exhibit hierarchical organization that mirrors the biochemical networks producing these compounds~\cite{zhou2018hyperbolic}. The cross-disciplinary validation of HypLLMs strengthens the theoretical foundation of hyperbolic geometry as a unifying framework for complex hierarchical systems. The remarkable success across diverse domains—detecting subtle patterns in brain networks, improving cancer outcome prediction, and enhancing multiscale reasoning in NLP—validates its ability to be a versatile tool across AI fields. In summary, hyperbolic approaches have demonstrated broad utility, powering improvements in natural language understanding, vision–language grounding, biomedical knowledge graphs, and even neuroscience.  This broad applicability suggests that hyperbolic principles may represent universal organizational patterns in both biological and artificial intelligence systems.

\section{Conclusion}
Hyperbolic geometry offers a compelling framework for advancing large language models, enabling them to capture hierarchical structures that are elusive in Euclidean space. This paper has provided a thorough discussion of hyperbolic large language models (HypLLMs), exploring their theoretical foundations, architectural approaches, and emerging applications. As demonstrated through our architectural taxonomy (\cref{tab:hypLLM_categorized}), and cross-domain applications (\Cref{fig:hypLLM_domains}), hyperbolic geometry provides a unifying mathematical framework that spans from hybrid integration approaches to fully curved-space architectures, with successful deployment across diverse fields including natural language processing, computer vision, multimodal representation learning, neuroscience, and biomedicine. We highlight the significant potential of hyperbolic geometry for enhancing the representation capabilities of large language models, particularly for data with inherent hierarchical structures.

Despite these advances, several open challenges remain. Looking ahead, we anticipate several transformative directions for HypLLMs. Hybrid-curvature architectures, such as mixture-of-curvature experts in models like HELM, and learnable-curvature architectures such as HiM, may unlock new capabilities by representing different aspects of data with optimal geometric spaces. 
Continued research on numerical stabilization, considering curvature-sensitive initialization and the development of unified benchmarks for hierarchical and multi-scale reasoning represent critical directions for future research.

\section*{Acknowledgments}
This work was supported by the DOE SEA-CROGS project (DE-SC0023191), AFOSR project (FA9550-24-1-0231). We also thank the computing resources provided by the High Performance Computing (HPC) facility at NJIT.

\newpage
\appendix
\section{Möbius Gyrovector Operations}
\label{sec:mobius}
The key operations include:

\textbf{Möbius addition:} For $\mathbf{x}, \mathbf{y} \in \mathcal{B}_r^n$,
\begin{equation}
\mathbf{x} \oplus \mathbf{y} = \frac{(1 + 2\langle\mathbf{x},\mathbf{y}\rangle + \|\mathbf{y}\|^2)\mathbf{x} + (1 - \|\mathbf{x}\|^2)\mathbf{y}}{1 + 2\langle\mathbf{x},\mathbf{y}\rangle + \|\mathbf{x}\|^2\|\mathbf{y}\|^2}.
\end{equation}

\textbf{Möbius scalar multiplication:}
\begin{equation}
r \otimes \mathbf{x} = \tanh(r \cdot \text{artanh}(\|\mathbf{x}\|)) \frac{\mathbf{x}}{\|\mathbf{x}\|}.
\end{equation}

\textbf{Möbius matrix-vector multiplication:} For matrix $\mathbf{M}$ and hyperbolic point $\mathbf{x}$,
\begin{equation}
\mathbf{M} \otimes \mathbf{x} = \text{exp}_0(\mathbf{M} \cdot \text{log}_0(\mathbf{x})).
\end{equation}

\textbf{Möbius matrix-vector multiplication.} For matrix $M \in \mathbb{R}^{n \times n}$ 
and hyperbolic point $x \in \mathcal{B}^n_r$,
\begin{equation}
M \otimes x = \exp_o(M \log_o(x)).
\end{equation}

\section{Hyperbolicity measure}
\label{sec:gromov-delta}
Gromov's $\delta$-hyperbolicity~\cite{gromov1987hyperbolic} quantifies tree-likeness, where $\delta \geq 0$ measures the deviation from perfect tree-like geometry. For any 4-tuple $(x, y, z, t)$ in a metric space, $\delta(x, y, z, t)$ is defined as half the difference between the biggest two of the following sums: 
\begin{equation}
d(x, y) + d(z, t), \quad d(x, z) + d(y, t), \quad d(x, t) + d(y, z).
\end{equation}

The space’s $\delta$-hyperbolicity is defined as the supremum of these values across all possible 4-tuples. Smaller $\delta$ values indicate a higher similarity to tree structures, making it more naturally suited for hierarchical modeling~\cite{nickel2017poincare, ganea2018hyperbolic}.

\setcounter{table}{2}
\renewcommand{\thetable}{\thesection.\arabic{table}}
\setcounter{table}{0}  

\section{HypLLM Models categorized by Task, Data Type, and Benchmarks}
\label{sec:organizehyp}
To summarize prominent hyperbolic LLMs (HypLLMs) architectures, we present \Cref{tab:hypLLM_models_task_data}, which organizes these models by their 
learning tasks, data modalities, and benchmark datasets.

\newpage

\begin{table*}[!ht]
\centering
\caption{Representative hyperbolic large language models (HypLLMs) with their learning tasks, dataset types, and benchmark datasets.}
\label{tab:hypLLM_models_task_data}
\fontsize{6.5pt}{8.5pt}\selectfont
\setlength{\tabcolsep}{2pt}
\renewcommand{\arraystretch}{1.35} 
\begin{tabularx}{\textwidth}{|p{2.6cm}|p{2.3cm}|p{2.0cm}|X|}
\hline
\textbf{Model} & \textbf{Task} & \textbf{Dataset type} & \textbf{Datasets} \\
\hline
Hyperbolic BERT~\cite{chen2024hyperbolic} 
& Entailment detection 
& Language 
& GAP~\cite{webster2018gap}, DPR~\cite{rahman2012resolving}, WSC~\cite{levesque2012winograd}, WG~\cite{rudinger2018gender}, PDP~\cite{davis2017first}, FIGER~\cite{ling2012fine}, Open Entity~\cite{choi2018ultra}, CoNLL-2003~\cite{tjong-kim-sang-de-meulder-2003-introduction}, Few-NERD~\cite{ding2021few}, TACRED~\cite{zhang2017tacred}, TACREV~\cite{alt2020tacred}, Re-TACRED~\cite{stoica2021re}, MRQA~\cite{fisch2019mrqa} \\
\hline
HiT~\cite{he2024language} 
& Mixed-hop prediction, multi-hop inference 
& Language 
& WordNet~\cite{miller1995wordnet}, FoodOn~\cite{dooley2018foodon}, DOID~\cite{schriml2012disease}, SNOMED~\cite{stearns2001snomed} \\
\hline
PoinCLIP~\cite{Srivastava2024} 
& Zero-shot image classification and retrieval 
& Image-text pairs 
& Food-101~\cite{bossard2014food101}, CIFAR-10~\cite{krizhevsky2009cifar}, CIFAR-100~\cite{krizhevsky2009cifar}, CUB-200-2011~\cite{WahCUB_200_2011}, SUN397~\cite{xiao2010sun}, Aircraft~\cite{Maji2013Aircraft}, DTD~\cite{cimpoi2014dtd}, Pets~\cite{parkhi2012pets}, Caltech-101~\cite{fei2004caltech101}, Flowers~\cite{Nilsback2008Flowers}, STL-10~\cite{Coates2011STL10}, EuroSAT~\cite{Helber2019EuroSAT}, RESISC45~\cite{Cheng2017RESISC45}, Country211~\cite{Radford2021CLIP}, MNIST~\cite{lecun2002gradient}, CLEVR~\cite{johnson2017clevr}, PCam~\cite{Veeling2018PCam}, SST2~\cite{Radford2021CLIP} \\ \hline

Hyperbolic BLIP-2~\cite{mandica2024hyperbolic} 
& Image captioning, image-text retrieval 
& Image-text pairs 
& MS COCO~\cite{lin2014microsoft} \\
\hline
HyperLLM~\cite{cheng2025large} 
& Recommendation 
& User-item tabular data 
& Amazon-Toys~\cite{mcauley2015image}, Amazon-Sports~\cite{mcauley2015image}, Amazon-Beauty~\cite{mcauley2015image} \\
\hline
HERec~\cite{ma2024harec} 
& Recommendation 
& User-item graph data
& Amazon-Books~\cite{ma2024harec}, Yelp~\cite{ma2024harec}, Google-Reviews~\cite{ma2024harec} \\
\hline
HyperSurv~\cite{xiongenhancing} 
& Survival prediction 
& Histological images-text pairs
& TCGA~\cite{tomczak2015tcga} \\
\hline
HySurvPred~\cite{yang2025hysurvpred} 
& Survival prediction 
& Histological images-genomics pairs
& TCGA~\cite{tomczak2015tcga} \\
\hline
ANTHEM~\cite{choudhary2022anthem} 
& Product search ranking 
& E-commerce search 
& E-commerce search data \\
\hline
HypLoRA~\cite{yang2024hyperbolic} 
& Arithmetic reasoning 
& Math problems 
& MAWPS~\cite{koncel2016mawps}, SVAMP~\cite{patel2021nlp}, GSM8K~\cite{cobbe2021training}, AQuA~\cite{ling2017program} \\
\hline
HoRA~\cite{yang2024enhancing} 
& Mathematical reasoning 
& Math problems 
& MAWPS~\cite{koncel2016mawps}, SVAMP~\cite{patel2021nlp}, GSM8K~\cite{cobbe2021training}, AQuA~\cite{ling2017program} \\
\hline
Hypformer~\cite{yang2024hypformer} 
& Graph node classification 
& Graph 
& Amazon2M~\cite{chiang2019cluster}, ogbn-proteins~\cite{hu2020open}, ogbn-arxiv~\cite{hu2020open}, ogbn-papers100M~\cite{hu2020open} \\
\hline
HELM~\cite{he2025helm} 
& Multiple-choice QA 
& QA benchmark 
& Wikipedia~\cite{wikimedia2024}, MMLU~\cite{hendrycks2020measuring}, ARC-C~\cite{clark2018think}, CommonsenseQA~\cite{talmor2018commonsenseqa}, HellaSwag~\cite{zellers2019hellaswag}, OpenBookQA~\cite{mihaylov2018can} \\
\hline
L-CLIP~\cite{he2025hypercore} 
& Image-text retrieval 
& Image-text pairs 
& CIFAR-10~\cite{krizhevsky2009cifar}, CIFAR-100~\cite{krizhevsky2009cifar}, ImageNet~\cite{deng2009imagenet}, RedCaps~\cite{desai2021redcaps} \\
\hline
HypGraphRAG~\cite{he2025hypercore} 
& Knowledge QA 
& Knowledge graphs 
& WebQSP~\cite{yih2016value} \\
\hline
HiM-Poincar\'e~\cite{patil2025hierarchical} 
& Mixed-hop prediction, multi-hop inference  
& Ontology 
& WordNet~\cite{miller1995wordnet}, FoodOn~\cite{dooley2018foodon}, DOID~\cite{schriml2012disease}, SNOMED~\cite{stearns2001snomed} \\
\hline
HiM-Lorentz~\cite{patil2025hierarchical} 
& Mixed-hop prediction, multi-hop inference 
& Ontology 
& WordNet~\cite{miller1995wordnet}, FoodOn~\cite{dooley2018foodon}, DOID~\cite{schriml2012disease}, SNOMED~\cite{stearns2001snomed} \\
\hline
SHMamba~\cite{yang2024shmamba} 
& Audio-Visual QA 
& Audio-video pairs 
& MUSIC-AVQA~\cite{li2022learning}, AVQA~\cite{yang2022avqa} \\
\hline
HMamba-Full~\cite{zhang2025hmamba} 
& Sequential recommendation
& Sequential interaction data 
& ML-1M~\cite{harper2015movielens}, Texas, California, New York \\
\hline
HMamba-Half~\cite{zhang2025hmamba} 
& Sequential recommendation 
& Sequential interaction data 
& ML-1M~\cite{harper2015movielens}, Texas, California, New York \\
\hline
\end{tabularx}
\end{table*}

\end{document}